\documentclass{river-journal}
\usepackage{rivps}
\usepackage{graphicx}
\usepackage{graphics}
\usepackage{subfigure}
\usepackage{cite}

\usepackage{amsmath,amssymb,amsfonts}
\usepackage{algorithmic}
\usepackage{textcomp}
\usepackage{xcolor}
\usepackage{hyperref}

\par

\begin{document}
\begin{opening}
\title{NL2CMD: An Updated Workflow for Natural Language to Bash Commands Translation}
\author{Quchen Fu, Zhongwei Teng, Marco Georgaklis, Jules White, Douglas C. Schmidt}
\institute{\textit{Dept. of Computer Science, Vanderbilt University
Nashville, TN, USA} \\
\texttt{quchen.fu@vanderbilt.edu, zhongwei.teng@vanderbilt.edu, marco.georgaklis@vanderbilt.edu, jules.white@vanderbilt.edu, d.schmidt@vanderbilt.edu}
}
\end{opening}


\subsection*{Abstract}
Translating natural language into Bash Commands is an emerging research field that has gained attention in recent years. Most efforts have focused on producing more accurate translation models. To the best of our knowledge, only two datasets are available, with one based on the other. Both datasets involve scraping through known data sources (through platforms like stack overflow, crowdsourcing, etc.) and hiring experts to validate and correct either the English text or Bash Commands. 

This paper provides two contributions to research on synthesizing Bash Commands from scratch. First, we describe a state-of-the-art translation model used to generate Bash Commands from the corresponding English text. Second, we introduce a new NL2CMD dataset that is automatically generated, involves minimal human intervention, and is over six times larger than prior datasets. Since the generation pipeline does not rely on existing Bash Commands, the distribution and types of commands can be custom adjusted. We evaluate the performance of ChatGPT on this task and discuss the potential of using it as a data generator. Our empirical results show how the scale and diversity of our dataset can offer unique opportunities for semantic parsing researchers. 


\section{Introduction}
Automating the conversion of natural language to executable computer programs is a long-coveted goal that has recently experienced a  resurgence of interest amongst researchers and practitioners. In particular, converting natural language to Bash (which is a shell scripting language for UNIX systems) has emerged as an area of interest, with the goal of automating repetitive tasks, such as file manipulation, search, and application-specific scripting. 

The NL2Bash problem can be described as a semantic parsing challenge, i.e., creating a mapping from natural language to a formal, executable representation~\cite{mooney2014semantic}. Significant efforts to tackle this problem have been sparked by the NLC2CMD competition at the NeurIPS 2020 conference. Our recent work in this competition yielded an architecture that improves the state-of-the-art performance in translating national language to Bash Commands from 13.8\% to 53.2\% \cite{fu2021transformer}. The transformer model that we created for this the NLC2CMD competition is currently the best-performing architecture for this problem~\cite{bharadwaj2021explainable}.

Until recently, the Bash translation problem has relied heavily on the availability of the NL2Bash dataset~\cite{lin2018nl2Bash}. This corpus of over 9,000 English-command pairs contains frequently used Bash Commands scraped from forums, tutorials, tech blogs, and course materials. Constructing the NL2Bash corpus involved hiring freelance software engineers, each assigned to manually search, browse, and enter data through a web interface. 

The freelancers working on this project constructed roughly 50 English language and Bash Command pairs per hour, prior to filtering and cleaning the dataset  \cite{lin2018nl2Bash}. This manual approach is clearly resource-intensive since it requires labor from specialized freelancers, which is time-consuming and expensive. Moreover, the approach is not scalable since the marginal cost of labor does not significantly diminish as the size of the dataset increases.

As commonly observed in this research domain, only marginal improvements in the accuracy of solutions to the NL2Bash problem have occurred since progress has been impeded due to the limited amount of annotated data. This paper extends our prior published work~\cite{9680023} on translating natural language to Bash Commands and provides the following new contributions beyond our prior work:
\begin{enumerate}
    \item It describes the use of a membership query synthesis technique to generate a large dataset of Bash Commands, expanding the available data to solve this problem,
    \item It demonstrates a back-translation technique that takes the generated Bash Commands and creates corresponding natural language pairs,
    \item It discusses a validation and verification technique for generated Bash Commands by converting them into an executable form and running them in an isolated environment that doubles as a data quality metric,
    \item It presents a new dataset called NLC2CMD, which is the largest dataset for translating natural language to Bash Command available to researchers and practitioners,
    \item It adopts a post-process addition to the original workflow, replacing placeholder values with actual arguments and making many of the translated commands executable in the Linux environment.
\end{enumerate}

Most importantly, our work suggests that the future of tackling this hard problem lies within the automation of Bash Command Synthesis and Back-translation to natural language representations. We have established a workflow that can be improved upon and used to maximize model performance with minimal labeling costs. Our approach has numerous advantages over prior work, with notable improvements in time efficiency, labeling costs, diversity of the dataset, and practicality.

The remainder of this paper is organized as follows: Section~\ref{overview} and~\ref{background} introduces the NLC2CMD problem and outlines recent developments in semantic parsing; Section~\ref{keyChallenge} summarizes the challenges for translating natural language to Bash Commands; Section~\ref{methodology} analyzes the performance of different model structures and training techniques; Section~\ref{corpus} describes our data generation/validation technique and statistics including data quality and comparison with existing datasets; Section~\ref{metricerr} discusses different metrics and error analysis for the state-of-the-art model on our new dataset; Section~\ref{rethink} offers an alternative approach using chatGPT for NLC2CMD; and Section ~\ref{conclusionfuturework} presents concluding remarks and outlines our future work.

\section{Research Problem Overview}
\label{overview}

Translating natural language into source code for software or scripts can help developers find ways to accomplish tasks in languages they are not familiar with, similar to how help forums like Stack Overflow are used today. As early as 1966, Sammet~\cite{Sammet1966TheUO} envisioned a future of automated code generation where people program in their native language. While generating software templates from configuration files is now common practice, research on translating natural language into code is still in a relatively early stage. 

Past research mainly focused on scripting languages or small code snippets. Various datasets have been created to aid research on generating code from natural languages. Examples of such datasets include WikiSQL for SQL~\cite{zhongSeq2SQL2017}, CoNaLa for Python~\cite{yin2018mining}, and NL2Bash for Bash~\cite{Lin2018NL2BashAC}.

This paper focuses on the task of translating natural language into Commands in the Bash scripting language. Translating natural language into Bash Commands is an example of semantic parsing, which means natural language is translated into logical forms that can be executed~\cite{Berant2013SemanticPO}. For example, the phrase ``how do I compress a directory into a bz2 file'' can be translated to the Bash command: \texttt{tar -cjf FILE\_NAME PATH}.

In the near term, natural language to Bash Commands translation is unlikely to replace discussion groups or help forums completely. They can, however, provide a quick reference mechanism that may improve on-demand code suggestions and popups generated by integrated development environments (IDEs). This type of AI-based approach complements other prior work, such as SOFix~\cite{Liu2018MiningSF}, which can fix bugs in code by mining postings in Stack Overflow. 

\section{Background and Related Work}
\label{background}
This section summarizes the background and related work surrounding the areas covered in this paper. Our contributions focus on advancing machine translation, where our approach is based on dataset synthesis. This research approach is novel in the domain of Bash Command translation.

\subsection{\bf Machine Translation}

Various architectures have been explored for different tasks of program synthesis from natural language. For example, Lin \textit{et al.}~\cite{Lin2017ProgramSF} achieved state-of-the-art generation of shell scripts using Recurrent Neural Networks (RNNs)~\cite{Mikolov2010RecurrentNN}. Likewise, Zeng \textit{et al.}~\cite{Zeng2020PhotonAR} utilized the Bert~\cite{Devlin2019BERTPO}-based encoder and a pointer-generator~\cite{See2017GetTT} decoder to generate SQL code from text. Moreover, ValueNet~\cite{Brunner2020ValueNetAN} (Transformer encoder + LSTM decoder with pointer networks~\cite{Vinyals2015PointerN}) was the first Text-to-SQL system incorporating values. In addition, Xu \textit{et al.}~\cite{Xu2020IncorporatingEK} improved upon the TranX~\cite{Yin2018TRANXAT} transition-based neural semantic parser to translate natural language into general programming languages, such as Python. 

The best results in prior work on the problem of translating natural language to Bash Commands were produced by Tellina~\cite{Lin2017ProgramSF}.  Tellina used the Gated Recurrent Unit (GRU) Network~\cite{Chung2014EmpiricalEO},  which is an RNN that achieved 13.8\% accuracy on the NLC2CMD metrics proposed by IBM~\cite{Agarwal2021NeurIPS2N}. The Tellina~\cite{Lin2017ProgramSF} paper produced the NL2Bash~\cite{Lin2018NL2BashAC} dataset and new semantic parsing methods that set the baseline for mapping English sentences to Bash Commands.

Transformer models generally have better accuracy and faster training times~\cite{Caswell2020AdvIGTrans} than RNNs~\cite{Mikolov2010RecurrentNN} on machine translation tasks. Prior research on machine translation has largely focused on the GRU architecture to translate natural language to Bash Commands. This paper enhances prior research by exploring the performance of several architectures on the NLC2CMD dataset. 

Our experiments with applying Transformer models to the natural language to Bash task show that they outperform other approaches, such as (1) RNNs that show an 18.4\% improvement and (2) Bidirectional RNN (BRNN) that show up to 4.4\% improvement~\cite{Schuster1997BidirectionalRN}.
Analyzing how model structural choices and prediction strategies affect model performance in natural language to command translation task~\cite{Agarwal2021NeurIPS2N} is thus a key contribution of this paper. Since the energy and accuracy metrics for model evaluation were specifically designed for the NL2CMD competition, potential improvements for the metrics are also discussed.

\subsection{\bf Dataset Synthesis}
Bash is a frequently-used command line scripting language. It thus offers a unique opportunity to generate diverse---and more importantly---executable commands more easily due to its relatively short and simple nature. To increase programmer productivity, the Bash Commands suggested by a tool should be both syntactically and semantically correct. If suggestions are not syntactically correct and cannot execute, programmers may simply ignore them since they distract from the task at hand. Moreover, if translations are not semantically correct, programmers may execute Bash Commands that do not accomplish the goal that they want to achieve, or worse, have negative impacts on the system (such as deleting important files or directories). 

Our work on Bash Command generation divides the synthesis into two steps: (1) scraping syntax and flag structures from the Bash manual pages for efficient command generation and (2) training a back-translation model for accurate command summarization. This approach enabled us to construct a dataset of English-command pairs that is over six times larger than the original NL2Bash dataset.

Bash manual pages give an introduction to Bash features and are \textit{the definitive reference on shell behavior}\cite{bashman}, providing complete and accurate guidance for Bash usage. Recent work has explored the use of manual page data for assistance in Bash to natural language translation ~\cite{bharadwaj-shevade-2021-explainable}, processing the page descriptions to aid the translation model. However, we found the manual pages offered additional insight and enough context into utility-flag relationships to generate an entirely new dataset from scratch.

Numerous approaches have incorporated dataset synthesis and augmentation in translation tasks. Nguyen \textit{et al.} \cite{nguyen2020data} explored the use of combining augmented data with the original dataset to boost the accuracy of neural machine translation between human languages. Zhao \textit{et al.} \cite{zhao-etal-2020-active} also explored data augmentation in neural machine translation to improve dataset diversification. Notably, Agarwal \textit{et al.} \cite{agarwal2021using} proposed using document similarity methods to create noisy parallel datasets of code, enabling the advancement of machine translation with monolingual datasets.

With dataset generation, transformer-based models~\cite{li2023text} have proven effective for parallel corpus mining in the domain of machine translation~\cite{DBLP:journals/corr/abs-2002-08155}. Previous research has tried using classification techniques, such as document similarity~\cite{agarwal2021using}, to identify translations from pre-existing corpora. 

\section{Key Research Challenges}
\label{keyChallenge}
This section summarizes key research challenges we encountered when translating natural language to Bash Commands and describes general obstacles the machine translation community is facing on these topics.

\subsection{\bf Challenge 1: Translating from an ambiguous language to precise Bash Commands is hard}
\label{ambiguous.section}
Translating human language into code is inherently hard. One reason is that human language is ambiguous by nature. As the famous Winograd test~\cite{Levesque2011TheWS} puts it, the sentence ``The trophy would not fit in the brown suitcase because \texttt{it} was too big'', \texttt{it} can either mean trophy or suitcase. While a human may be able to decide which one is correct, computers have a harder time since understanding this sentence requires ``the use of knowledge and commonsense reasoning''~\cite{Levesque2014OnOB}.

There are two general types of ambiguities~\cite{ambiguity}:
\begin{itemize}
    \item {\bf Genuine ambiguities}, where a sentence really can have two different meanings to an intelligent listener. An example of genuine ambiguity in the context of Bash Commands is ``merge file A with B in folder C''. This sentence has at least 2 interpretations: ``merge file A with B if B is in folder C'' or ``merge file A with B and put the result in folder C''.
    \item {\bf Computer ambiguities}, where the meaning is entirely clear to a listener, but a computer detects more than one meaning. A compute ambiguity can occur when multiple parse trees exist for a natural language sentence, such that when the tree is flattened the order of words for input can be undefined. 
\end{itemize}
Both types of ambiguities can affect the performance of translation from natural language to Bash Commands. 

\subsection{\bf Challenge 2: The natural language to Bash translation task is usually a many-to-many mapping}
\label{many2many.section}

Translation tasks are usually many-to-many mappings, which means there can be multiple correct translations for the same sentence. Moreover, even the sentence itself can have multiple methods of expression. As the size of the dictionary grows, there will be more possible translations for the same input. The process of creating the target sentences requires significant human effort.

Natural language is inherently flexible and Bash Commands can have functional overlap between different utilities. For example, when translating natural language to Bash the phrases \texttt{find the word "foo" in file "bar"} and \texttt{search in "bar" for "foo"} have the same meaning. Similarly, both \texttt{grep -w foo bar} and \texttt{cat bar $\vert$ grep -w foo} are valid translations.

\subsection{\bf Challenge 3: Paired English and Bash Commands data are not easily accessible}
\label{lowresource.section}

Machine translation models require many training examples. Collecting such a corpus is hard, however, especially for supervised learning that requires paired data (\textit{i.e.}, data with labels) an understanding of both the source and target languages is needed. Without a large number of training examples, therefore, it may be hard for the model to generalize beyond the small samples in the training set. 

Translating natural language to Bash Commands provides a unique challenge in which there are both a large number of English sentences and Bash Commands. However, paired data (\textit{i.e.}, English sentences with the corresponding Bash Commands) are not easily accessible. For paired sources, such as coding help forums like Stack Overflow, the question is usually a detailed description of the command that is summarized succinctly by humans. Writing Bash Commands requires considerable coding skills and is thus hard to crowd-source.

\subsection{\bf Challenge 4: Bash Commands change environments and are generally computer specific}
\label{lowresource2.section}

Bash Commands are often executed on the command line and are used for file manipulation, search, and application-specific scripting. When these commands are generated in large quantities they often result in deleted files, undefined behavior, and incredibly large searches. This output not only taxes the environment they run on if executed, but can damage the system itself by deleting critical files and directories.

In addition to potentially dangerous behavior, different file systems vary dramatically and humans use a variety of methodologies when organizing their file systems. This results in infinite unique file system configurations, each with different directory structures, permissions, and file names. With both generated Bash Commands and commands scraped from the Internet, a valid execution on one machine does not ensure a valid execution on another. Moreover, what may achieve the desired result on one machine may crash another.

\section{Research Questions}
\label{methodology}
\textbf{Which deep learning architectures perform best when translating between natural language and Bash Commands?} 

Since there is relatively little literature published on translating natural language to Bash Commands, an important concern is identifying which architectures published in other domains perform best. In particular, 
Sequence-to-Sequence~\cite{Sutskever2014SequenceTS} models have been studied extensively in the context of translations, so we explored their performance on this particular task. These models consist of two main components: an \emph{encoder} and a \emph{decoder}. The encoder turns the inputs into vectors and the decoder reverses the process. We compared different combinations of encoder-decoder layers, including RNN, BRNN, and Transformer, to translate the natural language to Bash Commands.

Chen \textit{et al.}~\cite{Chen2018TheBO} discovered that Transformer quality gains stemmed mostly from the Transformer encoder and that RNN decoders often have faster inference times. We therefore mixed and tested different combinations of encoder and decoder types. Table \ref{table:models} summarizes the performance comparison (measured in seconds) between different model structures. 
\begin{table}[bhtp]
\caption{\label{table:models} Model Performance Comparison}
\centering
{
\begin{tabular}{lllll}
\hline \textbf{Encoder} & \textbf{Decoder} & \textbf{Accuracy}& \textbf{Train} & \textbf{Inference}
\\\hline
Transformer & Transformer & $\mathbf{0.522}^{*}$ & 1625 & 0.126 \\
Transformer & RNN & - & - & - \\
RNN & Transformer & 0.486 & 1490 & 0.116 \\
RNN & RNN & 0.336 & $\mathbf{1151}^{*}$ & 0.069 \\
BRNN & Transformer & 0.495 & 1411 & 0.120 \\
BRNN & RNN & 0.476 & 1218 & $\mathbf{0.065 }^{*}$\\
\hline
\end{tabular}}
\end{table}

\begin{table*}[bhtp]
\caption{\label{table:leaderboard} The NLC2CMD Leaderboard}
\centering
\begin{tabular}{llllll}
\hline \textbf{Team} & \textbf{Model} & \textbf{Data Augment}& \textbf{Accuracy} & \textbf{Power} & \textbf{Latency}
\\\hline
Magnum & Transformer & No & $\mathbf{0 . 5 3 2}^{*}$ & 682.3 & 0.709 \\
Hubris & GPT-2 & No & 0.513 & 809.6 & 14.87 \\
Jb & Classifier+Transformer & Yes & 0.499 & 828.9 & 3.142 \\
AICore & Two-stage Transformer & No & 0.489 & $\mathbf{5 9 6 . 9}^{*}$ & 0.423 \\
Tellina~\cite{Lin2017ProgramSF} & BRNN (GRU) & No & 0.138 & 916.1 & 3.242 \\
\hline
\end{tabular}
\end{table*}

The results shown in Table~\ref{table:models} indicate that in this particular case, using the Transformer as both an encoder and decoder has the best accuracy\footnote{The OpenNMT~\cite{Klein2017OpenNMTOT} framework currently does not support a Transformer encoder + RNN decoder.}. Likewise, the model with an RNN as the decoder can reduce inference time by 50\%.

To provide a high-level perspective on how model architecture impacts performance, we analyzed the architectures of the top-performing teams in the NLC2CMD competition. Table~\ref{table:leaderboard} shows the Top 4 teams and the baseline model on the NLC2CMD Challenge leaderboard~\cite{Agarwal2021NeurIPS2N}. The Transformer architecture discussed in Section~\ref{arch.summary} was produced from an analysis of our team Magnum's architecture, which won the accuracy competition. 

AICore~\cite{Agarwal2021NeurIPS2N} won the energy track by having the least energy consumption with a two-stage prediction design consisting of two 2-layer Transformers. The first model predicted the template and the second model filled in the arguments. We suspect their small energy consumption is due to smaller model sizes (in contrast, the Magnum teams' model consisted of six layers). However, the gain in less energy consumption also came with a cost of lower accuracy (4.3\% decrease). 

Team Hubris~\cite{Agarwal2021NeurIPS2N} adopted a fine-tuned ensemble GPT-2 as the language model and achieved second place in accuracy. GPT-2 models are large (usually more than 5 GB) and power-hungry. It is therefore challenging to apply them as a background program running continuously in a terminal to suggest translations of Bash Commands. Another problem with GPT-2 ensembles is that their inference time (774M params) was prohibitive for real-world deployment, which requires fast response time and low energy consumption to run continuously in the background. Considerable effort is needed to compress and deploy GPT-2 ensembles to compete with other solutions.

Team Jb~\cite{Agarwal2021NeurIPS2N} augmented the training data using back-translation~\cite{Alzantot2018GeneratingNL} and created 78,000 augmented training samples. They also used the manual pages of Linux Bash Commands~\cite{Linux2018} to concatenate utilities with corresponding flags and generate an additional 200,000 new samples. Similar to Team AICore, they also used a two-stage model consisting of a classifier for utility prediction and a transformer for command generation. Interestingly, a large number of additional training samples was insufficient to overcome the architectural improvements of other teams. 

The results shown in Table~\ref{table:leaderboard} provide several key insights:
\begin{itemize}
\item Transformer models were the most popular choice. In this task, two-stage models performed worse than a single-stage-and-larger model.

\item GPT-2 approaches achieved near state-of-the-art accuracy, but produced much larger models compared to Transformers and had much longer inference times.

\item Data augmentation improved accuracy (Team Jb is 1\% more accurate than Team AICore) but had less impact than the model structure in this task (with the caveat that the two teams had similar---but not identical---models).
\end{itemize}
The experiments in the remainder of this paper use Transformer models since they were the best-performing architecture in the NLC2CMD competition.

\textbf{How do Bash Command parameters affect the performance of natural language to Bash translation?} 

As discussed in Section~\ref{lowresource.section}, obtaining training data of paired English and Bash Commands is hard. Without sufficient training data, the model may not be able to learn the entire vocabulary that it must translate to or from. Finding ways of reducing vocabulary size is thus essential to developing more accurate models.

Bash Commands typically consist of three terms: (1) utilities that specify the main goals of the command ({\em e.g.}, \texttt{ls}), (2) flags that provide metadata regarding command execution ({\em e.g.}, \texttt{-verbose}), and (3) parameters that specify directories, strings, or other values that the command should operate on ({\em e.g.}, /usr/bin). Each utility has a bounded number of flags that can be passed to it. In contrast, parameters have a much larger range of values. Training examples for translating natural language to Bash Commands provide values for the parameters, which can vary significantly between translated examples of the same command. 

We hypothesized that including the actual parameter values (such as \texttt{ls /usr/bin} and \texttt{ls /etc}) from the training examples would vastly increase the overall vocabulary size and decrease model accuracy. Our rationale for this hypothesis was that there were few paired examples of natural language and Bash Commands. Translation models therefore typically perform worse with large vocabulary sizes and limited training data. 

To test this hypothesis, we used the English and Bash tokenizers from the Tellina model~\cite{Lin2017ProgramSF} with our modification. 
\begin{figure}[hbt]
    \centering
    \includegraphics[width=0.6\textwidth]{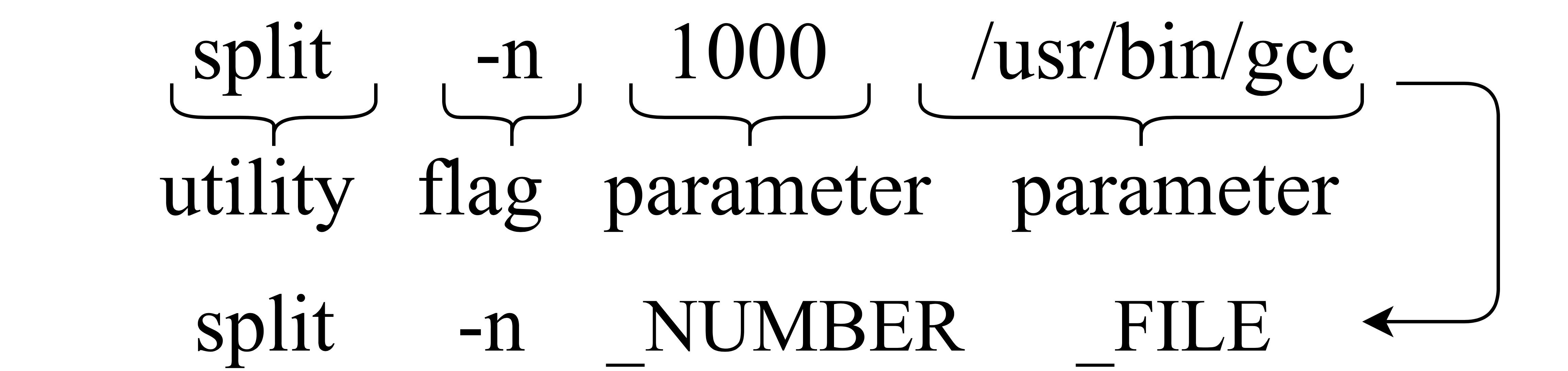}
    \caption{Example of a Bash Command}
    \label{fig:Bash_structure}
\end{figure}
As shown in Figure~\ref{fig:Bash_structure}, Bash tokens can be categorized as utilities, flags, and parameters (\textit{i.e.}, arguments, such as a specific path). The English tokenizer decapitalized all the letters and replaced parameters with generic forms. The Bash tokenizer parsed Commands into syntax trees with each element labeled as utility, flag, or parameter. 

Our accuracy metric focused mainly on the structure and syntactic correctness of the Bash Command. We therefore replaced all the parameters in Bash with their corresponding generic representations. For example, a folder path like \texttt{/usr/bin} is replaced with \texttt{PATH}. By applying this transformation, the Bash vocabulary size was reduced from 8,184 to 776 tokens, and the accuracy of the Transformer models we tested increased by 1.3\%. As shown in Table \ref{table:parameter}, we achieved accuracy and performance increases across all architectures, especially for the ones with less accuracy.

\begin{table}[bhtp]
\caption{\label{table:parameter} Parameter Replacement}
\centering
\begin{tabular}{llll}
\hline 
\textbf {Encoder}  & \textbf{Decoder} & \textbf{Accuracy} & \textbf {Accuracy (NP)}\\
\hline
Transformer & Transformer & 0.509 & \textbf {0.522*} \\
\hline
Transformer & RNN & - & - \\
\hline
RNN & Transformer & 0.448 & \textbf {0.486*} \\
\hline
RNN & RNN & 0.151 & \textbf {0.336*} \\
\hline
BRNN & Transformer & 0.483 & \textbf {0.495*} \\
\hline
BRNN & RNN & 0.301 & \textbf {0.476*} \\
\hline
\end{tabular}
\end{table}

\textbf{How to expand the amount of available Bash Command English language pairs without the hiring of external freelancers?} 

As discussed earlier, further innovation and developments when attempting to solve the natural language to Bash Command translation problem are severely restricted by the limited number of command-natural language pairs provided in the original dataset. This question not only deals with the most effective way of adding new and valid Bash Commands, but also deals with creating corresponding natural language pairs.

Scraping from online forums, such as StackOverflow, effectively gathers commands that could be added to the dataset, but which yield significant problems when dealing with invalid commands, duplicate commands, and commands of different programming languages. Another potential approach could be to augment the training data available, making minor changes to commands like the removal of a flag. However, this approach also poses challenges in differentiating between similar commands and adds minimal diversity of functions to the dataset.

We decided to create a Bash Command generator and use manual page data to synthesize entirely new Bash Commands. This approach allowed us to curate the synthesized dataset, maintaining similarities to the existing dataset, while still introducing informative data points.

To create corresponding natural-language components, we used a back-translation model with a transformer architecture. Transformer models have proven extremely effective with summarization tasks, which our back-translation was beginning to closely resemble.

\subsection{Summary of the Highest Performing Architecture} 
\label{arch.summary}

We tested several different data processing, architectural, and post-processing strategies, as discussed above. We now describe the best-performing model that we tested on the NLC2CMD competition data. Although this model will be improved by subsequent work, it provides a starting point for researchers focusing on natural language to Bash Command translation. In particular, our results show that the Transformer model is a robust foundation for future research in this area. 

Our Transformer model pipeline was built from the following six steps shown in Figure \ref{fig:pipeline} and described below:
\begin{enumerate}
  \item{\bf Parsers and filters} --
  The paired raw data first go through different parsers that convert English sentences and Bash Commands into syntax trees (data that cannot be parsed are removed).
  \item{\bf Flatten and pre-process} --
  The syntax trees are flattened and the parameters are replaced with their generic representations.
  \item{\bf Tokenizer} --
  The flattened sentence pairs are tokenized and dictionaries are created for English sentences and Bash Commands. 
  \item{\bf Transformer models} --
  Tokenized sentences are fed into Transformer models and Beam Searches are enabled to produce multiple translations.
  \item{\bf Ensemble} --
  The best-performing models on the validation dataset are chosen to create an ensemble.
  \item{\bf Post-process} --
  The translations produced by the ensemble model are post-processed by removing the placeholder arguments and inserting the values originally removed by the parser.
\end{enumerate}

\begin{figure*}
    \centering
    \includegraphics[width=0.8\textwidth]{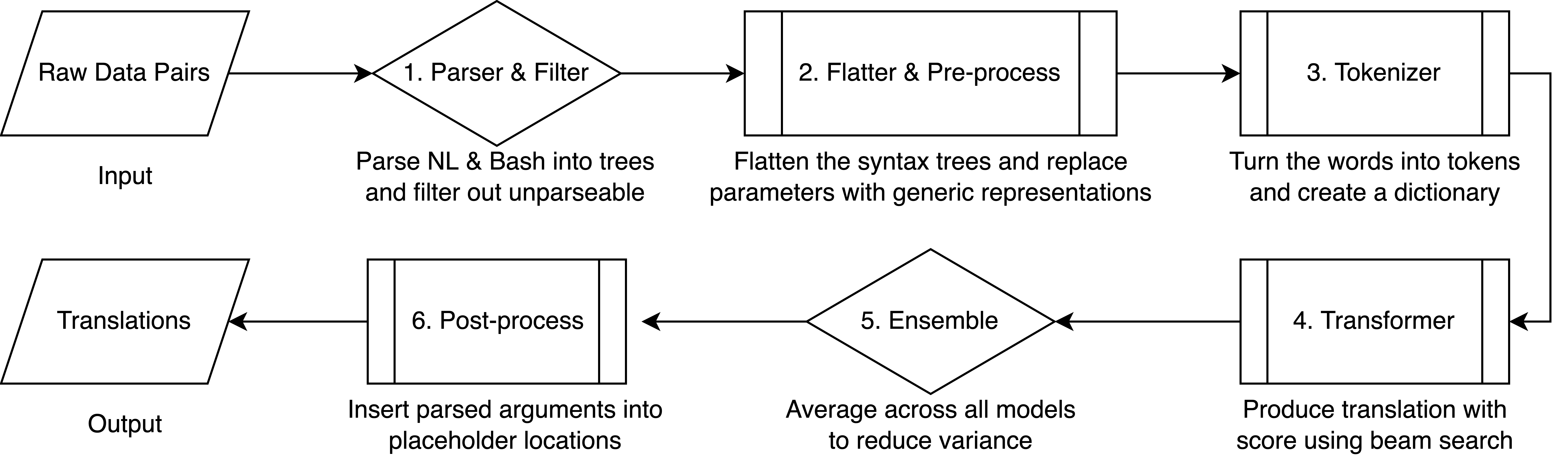}
    \caption{Pipeline of the NLC2CMD Workflow}
    \label{fig:pipeline3}
\end{figure*}

\subsection{Parsing and Tokenization} 
\label{arch.parsingandtokenization}
For our investigation, we used both the NLC2CMD dataset (which contains 10,347 pairs of English sentences and their corresponding Bash Commands) and our generated dataset (which consisted of 71,705 English sentence-Bash Command pairs). Of the 10,347 pairs of data in the original dataset, 29 had grammar issues and were therefore excluded. The size of this public dataset was relatively small in the natural language processing research field\footnote{In comparison, WMT-14 en-de (a popular dataset for machine translation benchmark) has 4.5 million sentence pairs.} and the goal for data processing was to create a small word vocabulary and utilize as much data as possible. Our generated dataset was significantly larger, so we shifted our focus to achieving higher quality commands, as opposed to using as many of the generated commands as possible.

\begin{figure}[hbpt]
\centering
\includegraphics[width=0.8\linewidth]{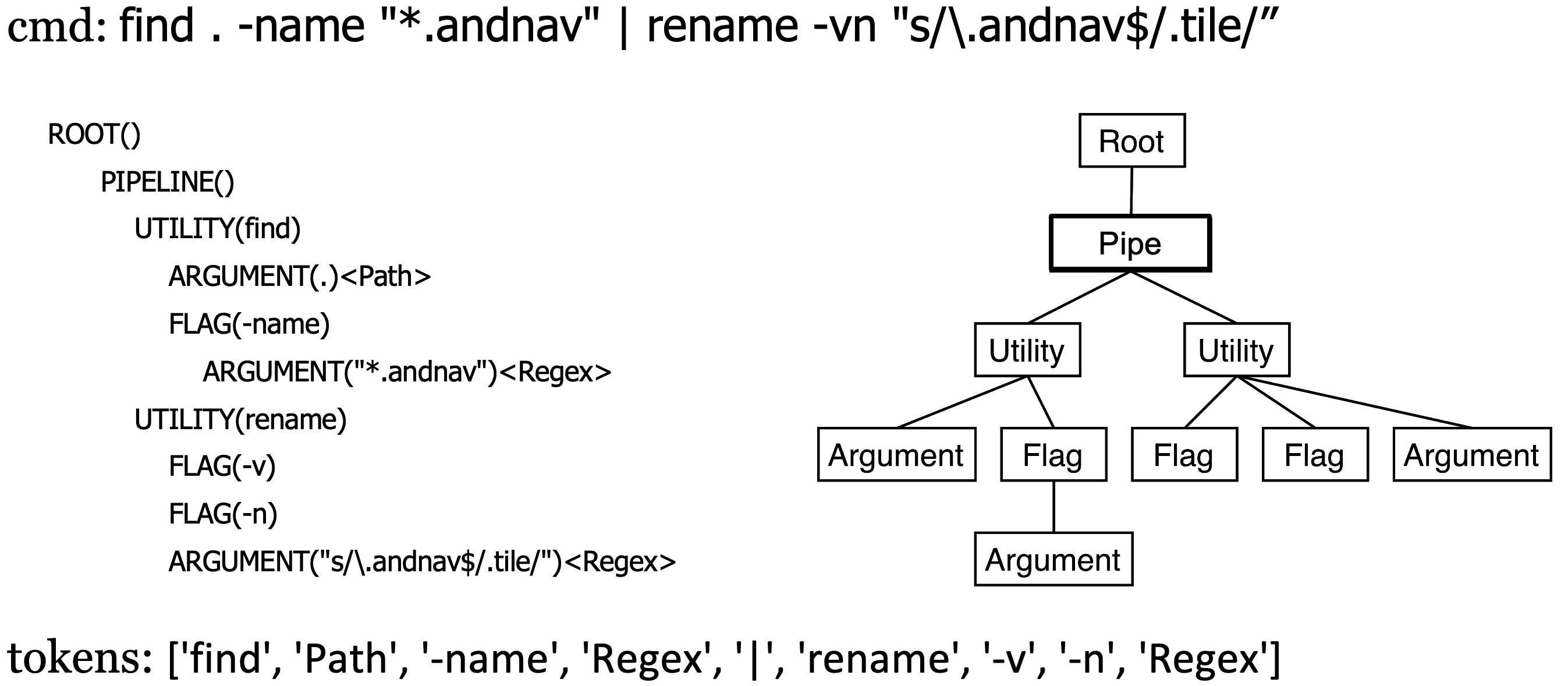}
\caption{Visualization of the Tokenization Process}
\label{fig:tokenization}
\end{figure}

Bash Commands can be complex and nested, as shown in Figure \ref{fig:tokenization}.
This structure helps explain why programmers may find it hard to create---or even comprehend---Bash Commands, thereby motivating the need for a customizable parser. Bash Commands can also be piped, which means the Commands may consist of multiple parts, with the output of the former part being the input for the latter one. 

We built our parser atop the Tellina~\cite{Lin2017ProgramSF} parser developed based on Bashlex~\cite{Bashlex} in prior work. This parser can parse a Bash Command into an abstract syntax tree (AST) composed of utility nodes, each of which may contain multiple corresponding flags and parameters. During the tokenization stage, utilities and flags are kept ``as is'' and parameters are categorized and replaced with \texttt{\_NUMBER, \_PATH, \_FILE, \_DIRECTORY, \_DATETIME, \_PERMISSION, \_TIMESPAN, \_SIZE}, with the default option of \texttt{\_REGEX}.

Natural language sentences are pre-processed by filtering out the stop words (e.g, ``a'', ``is'', ``the'', which carry little meaning). The remaining words are then decapitalized and lemmatized (preserving the common base form) to create a relatively smaller dictionary mapping. 

Our generator used a different parameter categorization strategy than the parser. The goal was to better align with the interpretation of the manual pages and the generation of high-quality commands. Parameter categorizations were similar, however, and the generator categorization was easily converted to the representation of the parser for training and inference with the transformer-based model.

\subsection{Model Details} 
\label{arch.modeldetails}

\begin{figure}[hbpt]
\centering
\includegraphics[width=0.7\linewidth]{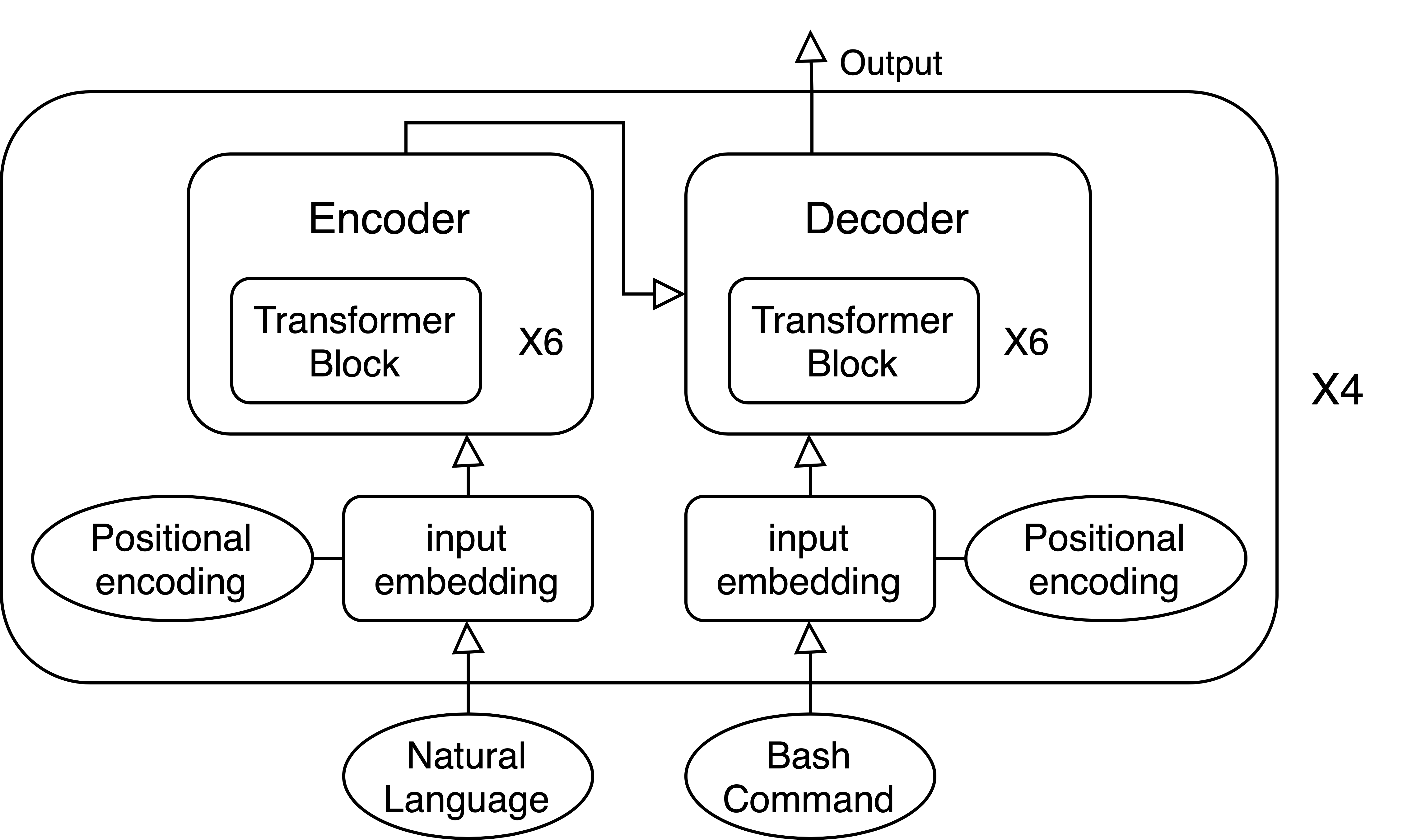}
\caption{Model Structure}
\label{fig:model}
\end{figure}

\begin{figure*}
    \centering
    \includegraphics[width=0.8\textwidth]{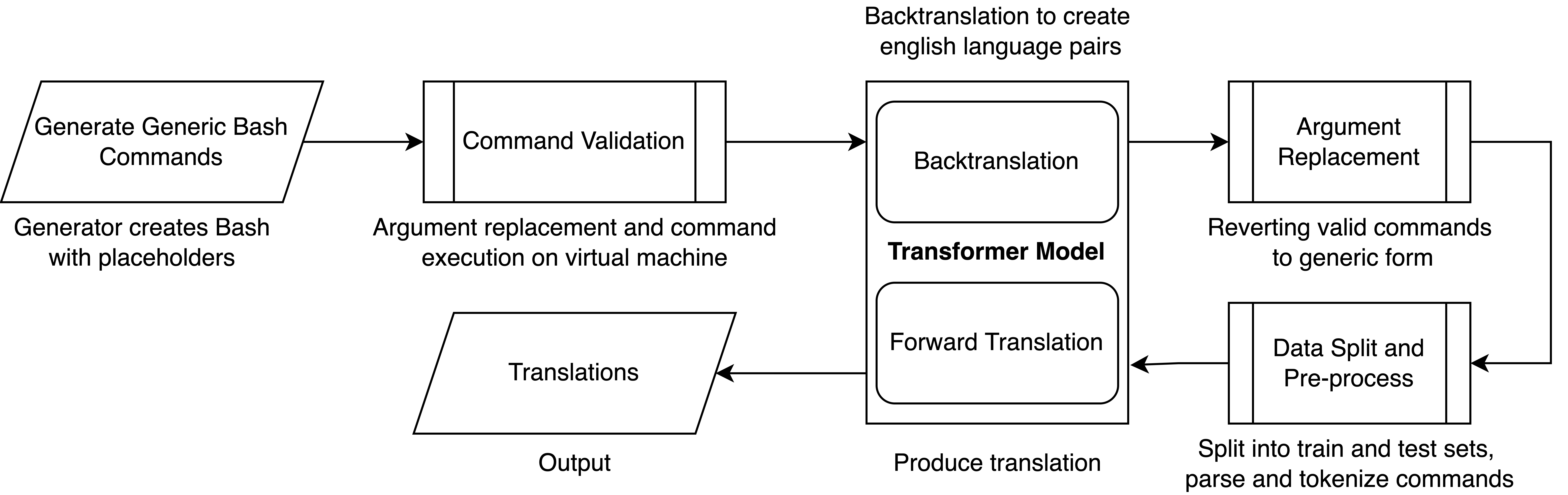}
    \caption{Updated Pipeline of the Dataset Generation and Translation}
    \label{fig:pipeline}
\end{figure*}

The model with the highest accuracy used a Transformer as both the encoder and the decoder, as shown in Figure \ref{fig:model}. 
The encoder and decoder each consisted of six layers. The model was trained for 2,500 steps and used an ensemble of the four top-performing single models.

The first positional weight was set to 1.0 and the rest of the weights were set to the exponential of beam scores capped by 0.5. We focused on training an efficient and robust model that can be deployed easily. The need to modify the network structure was therefore relatively low. Instead of FairSeq~\cite{Ott2019fairseqAF} (which allows users to modify the low-level network structure), we chose OpenNMT~\cite{Klein2017OpenNMTOT}, which is an open-source neural sequence learning framework to implement our Transformer model. 

We found that the Transformer model is sensitive to learning rate and larger batch sizes will produce better results. The detailed training hyperparameters are available on our GitHub repository~\cite{MagnumNLC2CMD2020}. Likewise, the guiding principle behind our tuning strategy is derived from Popel \textit{et al.}~\cite{Popel2018TrainingTF}. 

We trained our model on 2 Nvidia 2080 Ti Graphic cards with 64GB memory. Our model achieved 53.2\% accuracy on the hidden test dataset for the NLC2CMD competition and had the top performance in both inference time and energy consumption. We addressed challenge \ref{ambiguous.section} by masking out specific parameters. To limit ambiguity, the dataset itself also restricted the natural language description to a single sentence and the Bash Command to a single line~\cite{Lin2018NL2BashAC}. When the dataset was collected, the same Bash Command was paired with many English descriptions to increase language diversity~\cite{Lin2018NL2BashAC}, thereby addressing Challenge~\ref{lowresource.section}. 

\subsection{Post-processing}

The initial results of our model translation contained the placeholders inserted by the parser described above. This insertion was done to create a smaller dictionary mapping and improve the accuracy of the model. This result, although accurate, lacks practicality and many of the commands are confusing to humans and are far from executable. For example, the inclusion of {\_REGEX} in a command translated from the English language is vague and hard for anyone looking to use the workflow as a tool. To address this issue, we added post-processing of the translated commands into our workflow. Both inexperienced users and machines themselves can make use of executable commands, easily running them in a terminal, yet commands nested with vague placeholders are less useful. 

Using the parser from the pre-processing of the natural language, we extracted the parameters provided in the original corpus and used them to fill out placeholders in the created translation. Many translated commands had a perfect one-to-one replacement with the extracted parameters replacing all of the placeholders to create executable Bash Commands. Many other commands, however, contained more placeholders than parameters extracted from the parser. which resulted in partially replaced commands. Although these commands were not executable, they were also not implicitly incorrect translations.

For example, the description "remove all files in the current directory with a specific inode number" may have translated to \texttt{find . -inum Quantity exec rm {}}. In this translation, the specific inode number represented by the \texttt{Quantity} placeholder was not replaced as it never appeared in the original description. This demonstrates how the creation of partially replaced, non-executable commands can still be accurate translations, such that efforts towards partial replacement are worthwhile. This last step in the workflow increases usability and practicality surrounding the translation pipeline by converting Bash Command templates to executable or nearly executable commands that better resemble the purpose described in the natural language.

\section{Corpus Construction}
\label{corpus}

We were able to construct a corpus of ``Bash Command and natural language'' pairs six times larger than that of the original dataset. We created this large corpus by developing a generator to synthesize millions of Bash Commands, which were later validated and scaled. We then fed these commands through our back-translation model to create the corresponding natural language pairs.

\subsection{An Updated Pipeline}
\label{distribution}

In our updated pipeline we include the most successful aspects of our prior pipeline to generate an entirely new dataset and then train and test our top-performing model on our new dataset. Our state-of-the-art transformer-based model has proven effective in machine translation, so we decided to incorporate it both in our dataset generation and our training. Our updated pipeline, also demonstrated in Figure 6, consisted of the steps described below:

\begin{enumerate}
  \item{\bf Manual Page Scraping} --
  We scraped data from the manual pages to determine the syntax usage of 38 utilities. We also determined the flags associated with each utility and the categorization of the parameter, if any, associated with each of those flags.
  \item{\bf Generation} --
  With the syntactical structures, flags, and arguments for each utility, we generated over 1 million Bash Commands from different combinations of flags and piped commands. 
  \item{\bf Validation} --
  The generated commands were then replaced with actual arguments. For example, \texttt{[File]} was replaced with \texttt{temp.txt}. These commands were executed on a virtual machine and we discarded all commands that did not execute successfully with exit statuses of zero within a given time frame. 
  \item{\bf Scaling} -- 
  The validated commands were then converted into a form understandable by the parser, parsed, and scaled. This step involved preserving a similar proportion of commands with the \texttt{find} utility to the original dataset and ensuring there was a diversity of other utilities in the new dataset. Likewise, we discarded commands of over-represented utilities and commands that were parsed incorrectly by the parser.
  \item{\bf Back-translation} --
  The validated commands were then converted into a form understandable by the parser and fed to the back-translation model. This model was the same transformer-based model used on the original dataset, except trained in the reverse direction, using Bash Commands to predict natural-language sentences. This step created the corresponding natural language pairs for the generated dataset.
  \item{\bf Forward translation} --
  The new dataset was then split into training and testing and used to train and evaluate the model. For the validation, the best-performing models on the validation dataset were chosen to create an ensemble.
\end{enumerate}

\subsection{Bash Command Synthesis} 
\label{arch.Bashcommandgeneration}

Our generation stage involved scraping manual page information and assembling together commands from individual components. We used data gathered from the Linux manual pages to form syntax structures to help our generator understand the relationship between the different components from which commands are formed. Bash Commands generally consist of utilities followed by flags and arguments, although complexity can increase dramatically with the introduction of piped and nested commands. With sophisticated web-scraping techniques and some manual oversight, generated a mapping of utilities to their corresponding arguments and flags. Likewise, each flag had corresponding arguments, as the introduction of flags often increased the number of arguments in a command. 

We also scraped the syntax for each utility, creating templates to outline the context in which each utility is used and the order in which the flags and arguments appeared. In total, we created mappings and templates for 38 of the most common utilities appearing in the original NL2Bash dataset.

With the collection of this data, we generated thousands of commands with combinations of zero to three flags present in each command. The number of potential commands our generator is capable of producing is in the billions. However, we limited the number of commands produced for the following reasons:

\begin{itemize}
\item {\bf Quality preservation}. While generating commands that include large numbers of pipes and flags may yield many valid and executable commands, these commands are rare and thus do not commonly appear in the original training dataset. Other characteristics we tried to preserve from the original training dataset when synthesizing our own commands were general similarities in utility distributions and the ratio of commands with a pipe to those without.

\item {\bf Practicality}. In both of our validation and back-translation processes described below, the amount of time required to process the dataset scales with the number of commands in the dataset. After several hundred thousand commands, it becomes less practical to devote further resources to additional command generation.
\end{itemize}

In the scaling stage, we scaled the command generation to generally resemble the utility distribution of the original NL2Bash dataset. For example, the original dataset consisted of 63.44\% commands that began with the utility \texttt{find}, so we scaled the number of the \texttt{find} commands generated to represent a similar percentage of our generated commands. 

\begin{table}[bhtp]
\caption{\label{table:stats} Command Generation Process}
\centering
\begin{tabular}{llll}
\hline 
& \textbf {Generation}  & \textbf{Validation} & \textbf{Scaling}\\
\hline
Utility Count & 38 & 35 & 35 \\
\hline
Non-piped Commands & 570,436 & 60,926 & 38,557 \\
\hline
Single-piped Commands & 500,000 & 81,787 & 33,148 \\
\hline
\end{tabular}
\end{table}

We attempted to do this scaling for all generated utilities, although we achieved varying results. The original dataset consisted of 117 different utilities, while our generator only supported 38, as shown in Table~\ref{table:stats}. Moreover, many of these 38 utilities had poor or inconsistent documentation in the manual pages, making it hard to accurately collect all available flags and arguments and generate enough commands to match the distributions desired.

Another limiting factor in the distribution matching was the difficulty of generating valid commands for certain utilities. As described in the next section, every generated command was later validated to determine whether or not to include it in the dataset. The likelihood of commands of certain utilities being deemed invalid was significant. Generating large quantities of commands for those utilities therefore hindered their representation in the dataset.

Some utilities had few available flags and a large number of duplicate appearances in the original dataset. For example, the command \texttt{cd [Directory]} appeared 13 times in the training data as it originally appeared without placeholders. In the generated dataset, however, there were no duplicate commands, so the command only appeared once, thereby limiting the representation of that utility in the generated data.

The generated dataset not only included commands with a single utility but piped commands, as well. These commands consist of multiple utilities or commands that were concatenated, allowing the sharing of information during execution. Implementing support for piped commands involved analyzing the training data for common utility pairs piped together, generating commands independently for each utility, and joining them together with a pipe symbol. 

For example, the most popular utility pair in the training data was shown to be (\texttt{find},  \texttt{xargs}). In particular, of the piped commands, a command with the \texttt{find} utility was often followed by an \texttt{xargs} instruction piped afterward. Of the commands in the original training dataset, 31.36\% of them contained one or more pipes, with the mean number of pipes in each piped command being 1.45.

Due to the increased complexity of piped command support, we included some simplifications in the generation of piped commands. In particular, 70.33\% of the piped commands in the training data only contained a single pipe, so we only supported commands with one pipe. Moreover, in contrast to commands without pipes, the utility distribution matching for piped commands encompassed fewer utilities. 43.71\% of the single pipe commands in the training data consisted of \texttt{find} commands that were followed by \texttt{xargs}, \texttt{grep}, and \texttt{sort} commands, so we made the decision to support these combinations with piped command generation.

The inclusion of piped commands in the generated dataset dramatically increased the number of available commands to generate. The number of potential commands to generate scaled exponentially when pipes were introduced, yet this further pressured the validation process that executed all the generated commands sequentially. As a result, we limited the generation of piped commands to combinations from 500 different \texttt{find} commands concatenated to 1,000 different \texttt{xargs}, \texttt{grep}, and \texttt{sort} commands, totaling 500,000 commands synthesized in the first stage.

\subsection{Bash Command Validation}

To ensure the validity of the generated commands, each command was replaced with valid placeholder arguments and executed in an isolated environment. An example is \texttt{cd [Directory]} being converted to \texttt{cd abc} with the directory \texttt{abc} being available on the machine. To protect against undefined behavior, commands were executed in a virtual machine environment. 

The commands were executed in series and each exit status was measured by the program. The commands that were executed to completion and returned with exit statuses of zero were set aside, whereas those with non-zero exit statuses were discarded. The valid commands set aside were then converted back to generic commands, this time using the placeholders available in the original NL2Bash dataset. To prevent hanging, the commands validation script had a timeout of 0.5 seconds for each command before deeming it invalid. However, the percentage of commands deemed invalid due to timeouts was insignificantly small.

\begin{figure}[hbpt]
\centering
\includegraphics[width=0.5\linewidth]{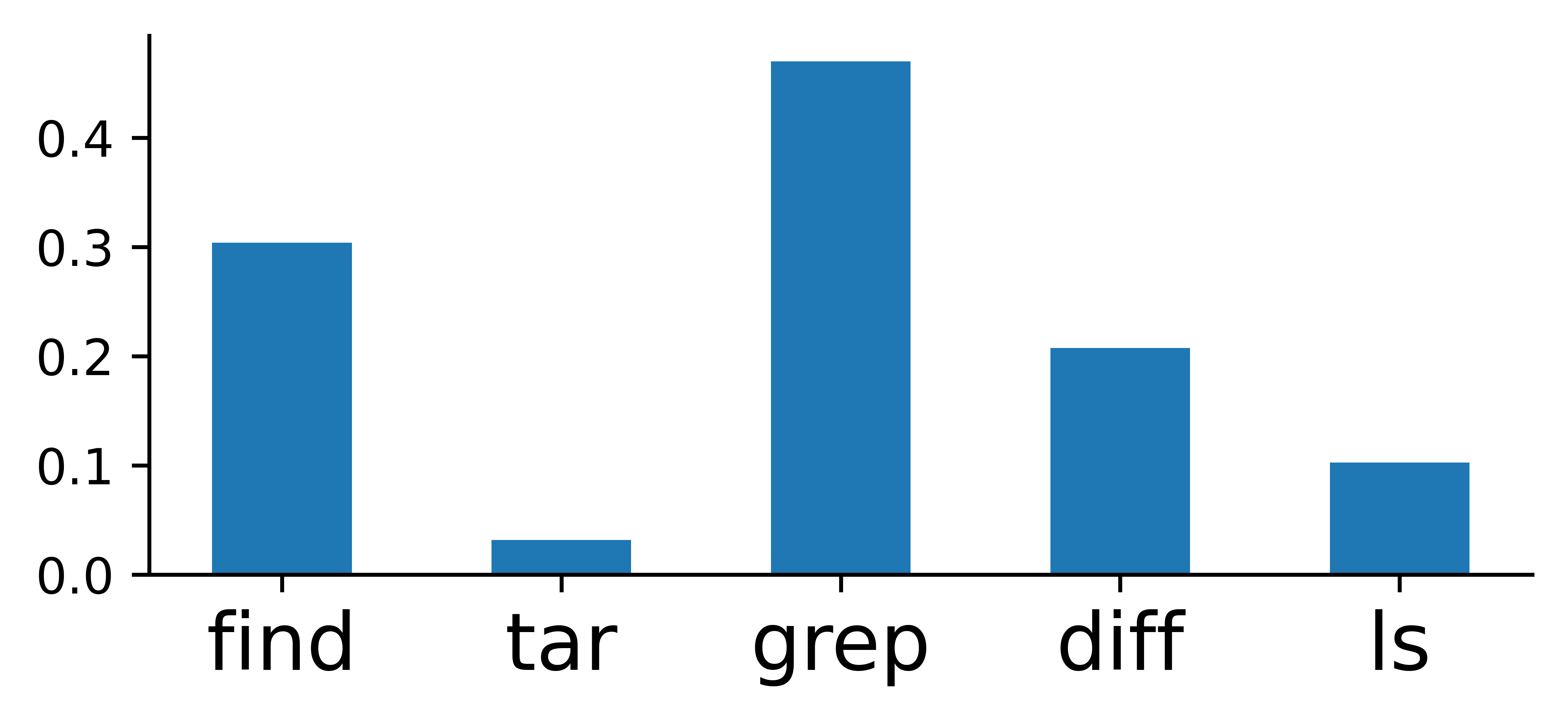}
\caption{Valid Command Rates For Generated Commands}
\label{fig:err}
\end{figure}

The percentage of commands that executed to completion with return statuses of zero varied dramatically for different utilities. Notably, commands like \texttt{grep} and \texttt{ls} were less prone to errors and approximately 50\% of them were able to complete execution with exit statuses of zero. \texttt{find}, the most common utility in the training data, had a validity rate of 30.4\%. Two utilities, \texttt{rev} and \texttt{rename}, had no generated commands that ran with exit codes of zero and were therefore removed from the generated dataset entirely. Overall, 13.3\% of the generated commands were deemed valid with validity rates of 10.7\% and 16.4\% for the non-piped and piped commands, respectively.

An exit status of zero does not guarantee a high-quality command. Commands can complete execution while still not achieving any change in the environment. For example, a directory change command like \texttt{cd .} might execute correctly, yet is a highly impractical command as it keeps the user in the same directory they started in.

Moreover, commands that fail the validation stage are not inherently incorrect. Generated commands are generic and placeholders are replaced for validation, so command failure can be a result of argument replacement. The virtual machine may not have the same folders, libraries, and files that are manipulated in a given command, which in some cases results in failure. While execution is an efficient way to discard many incorrectly generated commands, it does not correctly classify command validity in all cases.

\subsection{English Text Synthesis}

To generate the corresponding natural language for our generated Bash Commands, we used the same transformer-based model used in our original translation task. This model architecture proved extremely effective in machine translation tasks, so we reused this architecture for back-translation as opposed to another lower-performing model. In this case, we trained the model using data from the original NL2Bash dataset, but in the opposite direction, {\em i.e.}, attempting to predict natural language from Bash Commands. 

Once the model was trained, we used inference to predict the corresponding natural language for each command in the generated dataset. This completed the natural language component for every validated Bash Command and concluded the dataset generation process.

\subsection{Corpus Description and Statistics}

In total, our corpus contained 71,705 valid Bash Commands with corresponding English text. 69.9\% of these commands began with the \texttt{find} utility and the rest were distributed across 34 other utilities, which totals 35 utilities represented. The eight most popular utilities were \texttt{find}, \texttt{tar}, \texttt{grep}, \texttt{diff}, \texttt{ls}, \texttt{file}, \texttt{du}, and \texttt{cp}.

All our generated commands contained between zero and three flags for each utility within the command. Generally, commands without a pipe contained one utility, and commands with a pipe contained two, but in rare instances for specific utilities, like \texttt{xargs}, nested commands supported the inclusion of more utilities. As a result, the generated commands contained between one and four utilities, with the vast majority containing just one or two. Of the generated commands, 33,148 (46.22\%) of them contained a single pipe, while the rest contained zero pipes. Every  command generated was executed in a virtual machine command-line environment and was able to complete execution with an exit status of zero within 0.5 seconds.

\subsection{Data Quality}

To maximize the quality of the generated commands, we intentionally chose not to use the same argument-type placeholders as the parser when generating commands. As described above, the parser defaulted to classifying parameters as \texttt{\_REGEX}, which is a difficult placeholder for executable command generation since it can take many forms.

Instead, we used 15 argument-type placeholders, many similar to those from the parser, but generally centered around our goal of creating valid and executable commands. We wanted to choose argument-type placeholders that were as specific as possible, while still easily scraped and classified from manual pages in an automated fashion. As opposed to one argument type \texttt{\_REGEX}, we included \texttt{Pattern}, \texttt{FormattedString}, and \texttt{Separator} to differentiate between the different types of regular expressions and improve the particularity of the generated commands. This decision was possible because the manual pages also differentiated between these argument types, so classifying these flags with these corresponding argument types based on keywords in the manual pages required no significant extra work.

Although our strategy for dealing with parameters and placeholders differed at the command generation stage, we did not want this inconsistency to propagate to our translation. The parser and its corresponding placeholders were shown effective in preprocessing, so after command generation, we converted the placeholders to match those of the parser to include in our synthesized dataset. This conversion was generally straightforward, with each of our chosen placeholders mapped to only one parser placeholder.

Moreover, our validation of all the generated commands through the injection of actual arguments into the placeholders and execution in a virtual machine environment ensured a high level of quality for our generated commands. Of our 570,476 no-pipe commands generated, 60,926 commands  (10.7\%) of our generated commands were executed to completion with exit statuses of zero.

This level of validation was not performed on the initial dataset.  We found that many of the commands provided in the original NL2Bash dataset were unable to execute to completion with exit statuses of zero in our environment. This result suggests the possibility of Bash Command generation eventually yielding a higher percentage of valid commands than those manually verified by expert freelancers.

\subsection{Comparison to Existing Dataset} 

As mentioned in Section~\ref{arch.Bashcommandgeneration}, we attempted to match the distribution of the training data when determining the ratio in which we generated commands for different utilities. Commands with the \texttt{find} utility amounted to the majority of the training commands and were the limiting factor when determining the size of the generated dataset. Despite these efforts, there were key differences between the existing and generated dataset, as shown in Table~\ref{table:leaderboard2}.
\begin{table}[bhtp]
\caption{\label{table:leaderboard2} New vs. Existing Dataset}
\centering
\begin{tabular}{llllll}
\hline \textbf{Category} & \textbf{Original} & \textbf{Generated (Raw)} &\textbf{Generated (Valid)}
\\\hline
Total Commands & 10,348 & 1,070,436 & 71,705 \\
Piped Commands & 3246 & 500,000 & 55,931 \\
Distinct Utilities & 117 & 36 & 36 \\
\hline
\end{tabular}
\end{table}

\begin{figure}[hbpt]
\centering
\includegraphics[width=0.5\linewidth]{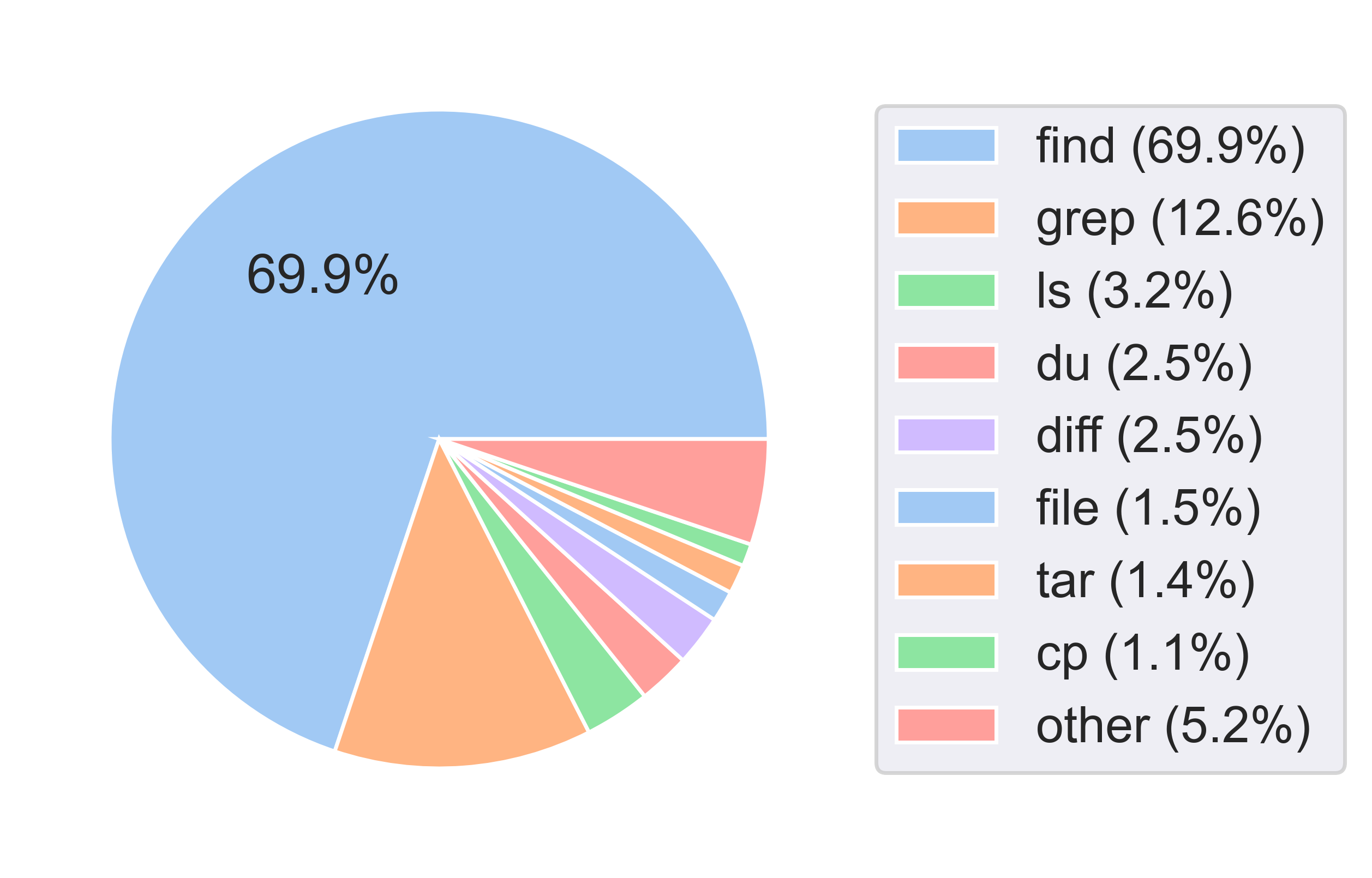}
\caption{Utility Distribution in the Generated Dataset}
\label{fignew}
\end{figure}

\begin{figure}[hbpt]
\centering
\includegraphics[width=0.5\linewidth]{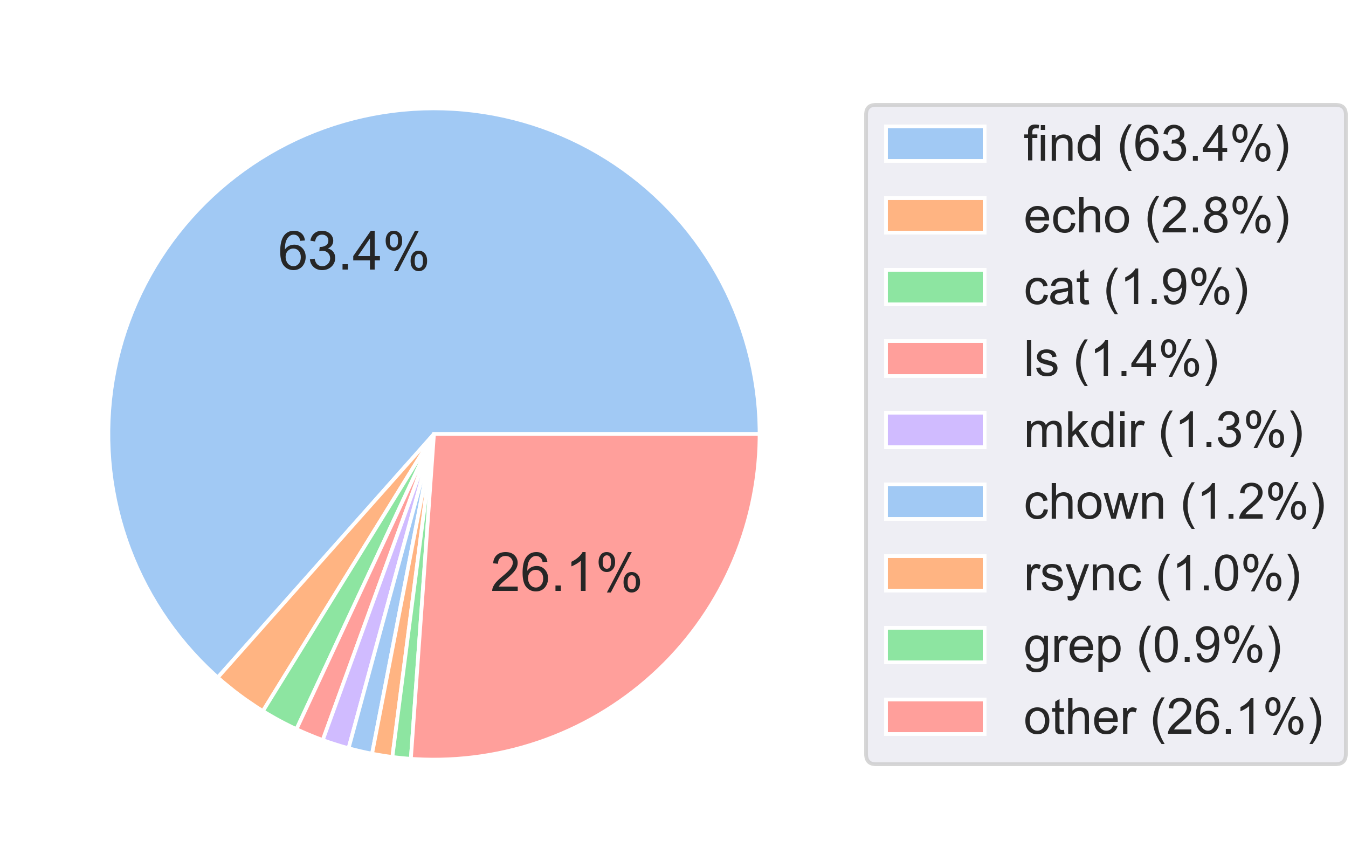}
\caption{Utility Distribution in the Original NL2Bash Dataset}
\label{figori}
\end{figure}

The original dataset contained a large diversity of commands, with 117 different utilities present. Since our generator only supported 38 utilities, many utilities represented a larger proportion of the dataset to account for those that were missing. For most of the included utilities, the quantity of generated commands for a given utility drastically outnumbered the number of commands for the utility in the original dataset. These utility distributions of the two datasets are shown in Fig~\ref{fignew} and Fig~\ref{figori}, where the most common utilities in the generated dataset are plotted alongside those of the original dataset.

There are several reasons for the differences in the composition of the two datasets. For example, the existing dataset contained many duplicate commands or commands of the same structure applied to different files, directories, or other arguments. In contrast, this level of duplication did not occur for the synthesized dataset since every Bash Command generated is unique. This duplication also resulted in popular commands with limited use cases, such as \texttt{cd [Directory]} being underrepresented in our generated commands with respect to the existing dataset. 

More generally, the original dataset contained large quantities of popular commands and underrepresented more obscure commands or those with unpopular flags due to the data-gathering strategy of searching online forums, as shown in Fig~\ref{figori}. 
This strategy resulted in a heavily biased dataset and little-to-no coverage of some powerful---yet uncommon---flags. Our generated dataset gave no preference to popularity and treated all flags equally, although those less likely to result in errors or time-consuming execution times were often filtered in validation stages. As a result, our dataset was significantly more diverse, with many unusual flag combinations represented that were not in the original dataset.

Moreover, our synthesized dataset included up to three flags for each utility in the command and at most a single pipe within the command. This configuration meant that every generated command had a maximum of two utilities and six flags, excluding a small minority of nested commands. Although the majority of the commands in the existing dataset matched this demographic, many commands in the training data had several pipes or contained more than three flags proceeding a utility. This configuration meant that although the generated dataset included a large number of diverse commands, the commands were generally shorter in length and minimally complex.

The last key difference involved the validity of the commands. Since the entirety of the generated dataset went through the validation process and only the commands deemed valid were kept, 100\% of the resulting dataset was valid. When the original dataset was run under the same conditions, only 2,360 commands were able to execute with exit statuses of zero within the 0.5-second time frame, for a validity rate of 22.8\%.

In summary, the analysis above shows that although the two datasets were similar in many ways including the utility composition, there were key differences in utility and flag diversity, as well as validity rates.

\section{Metrics and Error Analysis}\label{metrics}	
\label{metricerr}
This section discusses different metrics and performances of the transformer-based Magnum model on both datasets. Section \ref{metricerr.metric} describes the accuracy metric and proposes an improved energy metric, Section~\ref{section.result} summarizes the accuracy performance of the Magnum model on the new NL2CMD dataset, and Section \ref{metricerr.errror} analyses the distribution of different error types on the original dataset.

\subsection{Metrics} 
Below we describe the accuracy metric and propose an improved energy metric.

\label{metricerr.metric}
\textbf{Accuracy:} The ideal metric for an evaluation would check if the predicted Bash Command produces the same result as the reference answer. That metric is not practical, however, since establishing a simulated environment for 10K variant situations is beyond the scope of this paper. Instead, our scoring mechanism specifically checks for structural and syntactic correctness that ``incentivizes precision and recall of the correct utility and its flags, weighted by the reported confidence''~\cite{Agarwal2021NeurIPS2N}. The metric first defines two terms: Flag score ${S_{F}^{i}}$ and Utility score $S_{U}^{i}$. 

As shown in Equation \ref{eq:flag}~\cite{Agarwal2021NeurIPS2N}, the flag score is defined as twice the union of reference flags and predicted flags number minus the intersection, scaled by the max number of either reference flags or predicted flags. 
\begin{equation}
S_{F}^{i} (F_{\text{pred}}, F_{\text{ref}}) = \frac{1}{N} \Big( 2 \times |F_{\text{pred}} \cap F_{\text{ref}}| - |F_{\text{pred}} \cup F_{\text{ref}}| \Big)\label{eq:flag}
\end{equation}
The range of flag scores is between -1 and 1. 

As shown in Equation \ref{eq:utility}~\cite{Agarwal2021NeurIPS2N}, the utility score is defined as the number of correct reference utilities scaled by capping flag score between 0 and 1, minus the number of wrong utilities, scaled by the max number of either reference utilities or predicted utilities.
\begin{small}
\begin{equation}
S_{U} = \sum_{i \in [1, T]} \frac{1}{T} \times \bigg( |U_{\text{pred}} = U_{\text{ref}}| \times \frac{1}{2} \Big(1 + S_{F}^{i} \Big) - |U_{\text{pred}} \not= U_{\text{ref}} | \bigg)\label{eq:utility}
\end{equation}
\end{small}
By summing all the utility scores within a predicted command, the range of normalized utility scores is between -1 and 1.

\textbf{Energy:} The measurement and reporting of energy consumption of natural language programming (NPL) models is a relatively new phenomenon~\cite{Strubell2019EnergyAP}~\cite{Cao2020TowardsAA}. As Henderson \textit{et al.}~\cite{henderson2020systematic} pointed out, part of the reason stems from the complexities of collecting the result. In particular, according to Appendix B of Henderson \textit{et al.}~\cite{henderson2020systematic}, out of 100 NeurIPS papers from the 2019 proceedings, only 1 measured energy consumption in some way, whereas 45 measured runtime performance. 

To address this gap, the NeurIPS 2020 conference recommended ``energy'' as a more direct way of measuring environmental impact.	We found the current energy metric used by the NL2CMD competition was not ideal, however, since it used estimated attributable power draw (mWatts) to compute scores. This metric disproportionately punished models with less inference time.	

For example, the GPT-2 model with an inference time of 14.87 seconds should have consumed a huge amount of energy (considering the model size). On the leaderboard shown in Table~\ref{table:leaderboard} the power metric is even less than the baseline, which is a much smaller model (GRU) and the inference is 3.24 seconds. Moreover, energy$_{mWh}$ can be easily affected by trivially extending inference time. For example, by simply sleeping 3 seconds after each batch, the performance of a test submission can be improved from 682 to 88 on the leaderboard. A potential fix would be to measure the total energy consumed instead of the power since it punishes both bigger model sizes and longer inference. 

\textbf{Validity:} Another metric measure the quality of Commands that datasets used. Our validity rate metric measured the percentage of commands that were able to execute to completion with exit statuses of zero within a 0.5-second time frame when replaced with standard replacement values. Although this metric is not entirely reliable due to the complex nature of computers behaving differently and containing different file systems, it provided valuable insight into the general correctness of the commands in the dataset. 

Commands that {\em passed} the validity test demonstrated a baseline level of error aversion and proved they did not result in undefined behavior under the given conditions. Commands that did {\em not pass} the validity test was not implicitly incorrect, however, as they could have executed correctly on a different machine or simply required longer than the allotted 0.5 seconds to execute. 

\subsection{Synthesized Dataset Results}
\label{section.result}

We split our generated dataset 80/20 into training and testing sets, respectively, before training our transformer-based model on the training dataset. We then evaluated our results and achieved an accuracy score of 31.63\%, as shown in Table~\ref{table:modelperf_data}.

\begin{table}[tbph]
\begin{center}
 \caption{\label{table:modelperf_data} Comparison of Model Performance Across Datasets}
\begin{tabular}{|c|c|c|}
\hline
\textbf{Training Dataset}  & \textbf{Test Dataset} & \textbf{Accuracy}\\
\hline
NL2Bash & NL2Bash & 52.3\%\\
\hline
NLC2CMD & NLC2CMD & 31.6\%\\
\hline
NLC2CMD & NL2Bash & -13\%\\
\hline
NL2Bash & NLC2CMD & -6.67\%\\
\hline
NLC2CMD+NL2Bash & NL2Bash & 48\%\\
\hline
NLC2CMD+NL2Bash & NLC2CMD & 31.7\%\\
\hline
\end{tabular}
\end{center}
\vspace{-0.2in}
\end{table}

This relatively low score demonstrates the difficulty of the dataset. We conjecture this result occurred because the dataset contained utility-flag combinations that were not present in the original dataset. These combinations resulted in inaccurate back-translations for the natural language components, which made predictions for the model extremely hard.

We also suspect over-fitting for the model trained on the NL2Bash dataset due to its relatively small size. We found that the model cross-trained on both datasets demonstrated good performance on both test sets, while the model trained on a single dataset performed poorly on the test set of another dataset. This result again highlights the importance of having a large and diverse dataset for increasing model robustness.

\subsection{Error Analysis} 
\label{metricerr.errror}

Previous research~\cite{Lin2018NL2BashAC} listed the top three causes of incorrect predictions as sparse training data, utility errors, and flag errors. Since sparse training data is a subjective metric, we only analyzed the incorrect utility and flag predictions. We used a separate, independently-created testing dataset of 1,867 samples (previous work manually analyzed 100 samples from the dev dataset collected the same way as the original training dataset) from the original training dataset and evaluated the accuracy results in more detail. 

\begin{figure}[hbpt]
\centering
\includegraphics[width=0.4\linewidth]{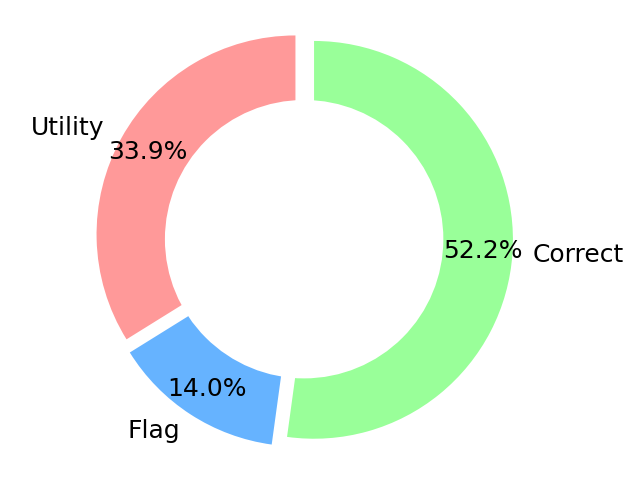}
\caption{Percentage of Utility and Flag Errors on Original Dataset}
\label{fig:err2}
\end{figure}

Figure \ref{fig:err2} shows that over two-thirds of all errors are utility errors, so the variety of flags is less significant than having enough data for each utility.
This result shows why our new dataset resulted in a significantly lower accuracy score. In particular, utilities that are less common in the original dataset have larger quantities of commands.

\begin{figure}[hbpt]
\centering
\includegraphics[width=0.7\linewidth]{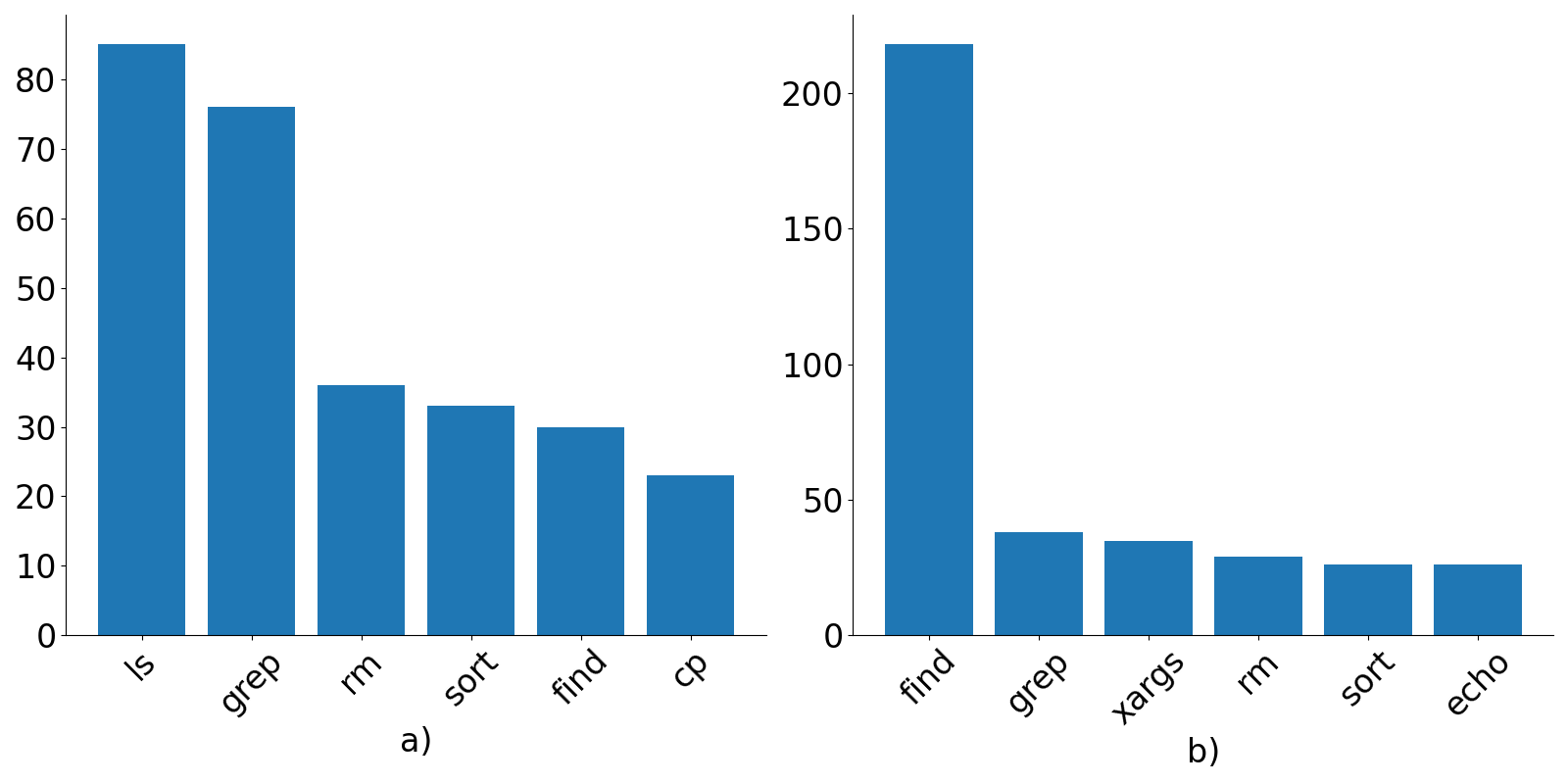}
\caption{(a) Distribution of Reference Utilities that are Wrongly Predicted. (b) Distribution of Wrongly Predicted Utilities}
\label{fig:utility}
\end{figure}

Figure \ref{fig:utility} shows that among the top six incorrectly predicted utilities, \texttt{ls} and \texttt{grep} are the most frequently confused with \texttt{find}. 
This confusion was expected since the functionality of these three utilities overlapped significantly and were among the most frequently used Bash Commands. By manually examining the incorrect predictions, we also found that these three utilities appear in many piped commands, which helps explain the large proportion they comprised in all the incorrectly predicted utilities. Our synthesized dataset contained both larger absolute and relative quantities of piped commands, so this result further underscored the difficulty of the model in predicting them correctly.

\section{Rethinking NL2CMD in the age of ChatGPT}
\label{rethink}
There is a widespread belief among experts that the field of natural language processing (NLP) is currently experiencing a paradigm shift~\cite{liu2023pre} as a result of the introduction of LLM (Large Language Models)~\cite{bommasani2021opportunities}, with chatGPT~\cite{bang2023multitask} being the leading example of this new technology. With this new technology, many tasks that previously relied on fine-tuning pre-trained models can now be achieved through prompt engineering~\cite{white2023prompt}~\cite{white2023chatgpt}, which involves identifying the appropriate instructions to direct the language model (LLM) for specific tasks. To evaluate the effectiveness of chatGPT, we conducted tests on the original NL2BASH dataset, and the results were exceptional. Specifically, we found that chatGPT achieved an accuracy score of 80.6\% on the test set under zero-shot conditions. Although there are concerns about the possibility of data leakage in LLM-based translation due to the vast amount of internet text in the pre-training data, we have confidence in the performance of chatGPT, given its consistent ability to achieve scores of 80\% or higher across all training, testing, and evaluation datasets.

\begin{figure*}
    \centering
    \includegraphics[width=0.6\textwidth]{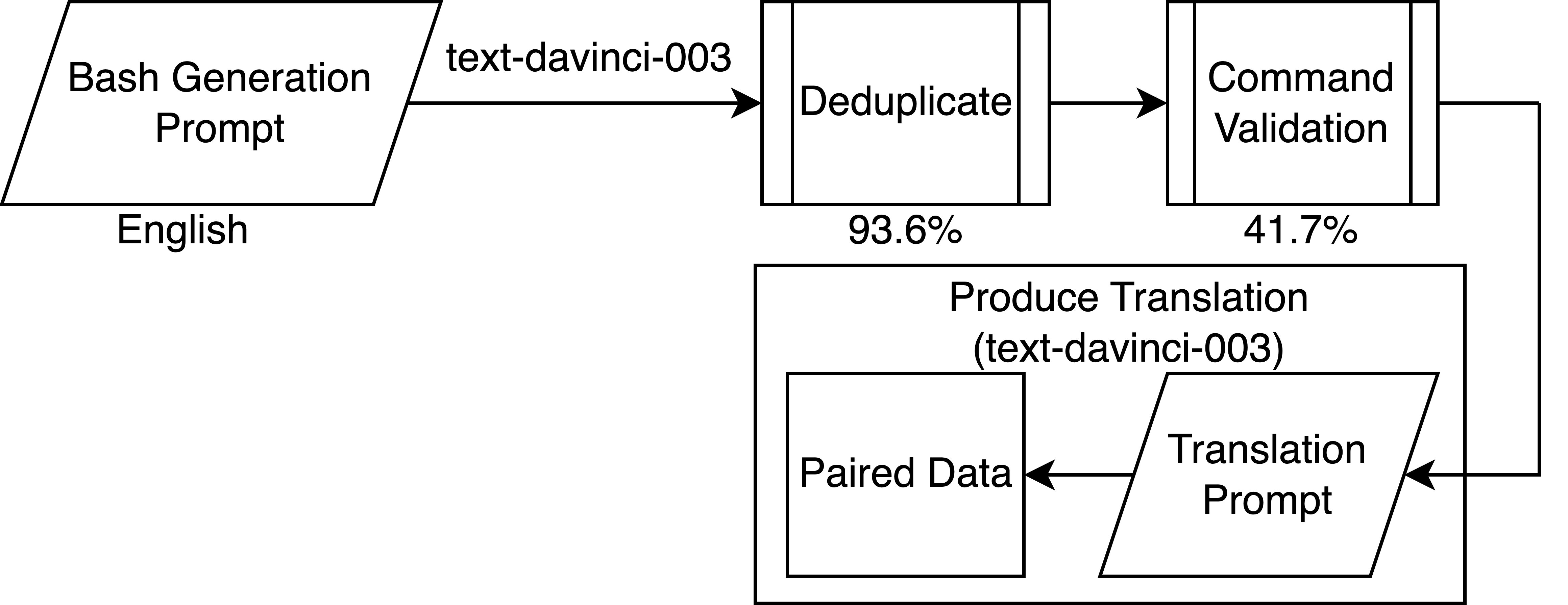}
    \caption{Streamlined Pipeline of the Dataset Generation and Translation}
    \label{fig:pipeline2}
\end{figure*}

We have conducted further exploration into the potential of streamlining our data generation pipeline with the assistance of ChatGPT, as shown in Figure~\ref{fig:pipeline2}. In order to generate Bash commands, we utilized the prompt \textit{Generate bash command and do not include example}. We set the "temperature" parameter to 1 for maximum variability. These generated commands were then subjected to a de-duplication script, resulting in a surprisingly low duplicate rate of 6\% despite prompting the system 44671 times. Subsequently, the data were validated using the same bash parsing tool previously mentioned, and 41.7\% of the generated bash commands were deemed valid. The preprocessed bash commands were combined with the prompt \textit{Translate to English}, yielding a paired English-Bash dataset with a size of 17050. We set the \textit{temperature} parameter to 0 for reproducibility.

In order to assess the quality of this generated dataset, we tested the performance of augmenting the original dataset with the generated version and found no performance drop. We further tested this approach by setting the \textit{temperature} parameter to 1 to introduce more variability, which yielded different English sentences for each Bash command, serving as a useful data augmentation tool.

This suggests that the ChatGPT-generated dataset is of higher quality than our previous pipeline. Furthermore, the performance of training on generated data and evaluating on NL2Bash was greatly improved, with the score increasing from -13\% to approximately 10\%. It is important to note that this is only a preliminary exploration into using ChatGPT as a data generation tool, and our observations represent a lower bound on the potential benefits of this method.

What is particularly groundbreaking about this approach is the efficiency with which it was implemented. Whereas the previous pipeline took two months to build, the ChatGPT streamlined version was completed in just three days. We have made our code and dataset available on Github~\cite{BashGen2022}. Notably, the distribution of generated utilities displayed a much smaller long tail effect~\cite{shen2023data}, suggesting that it more accurately captures the command usage distribution.

\section{Concluding Remarks}
\label{conclusionfuturework}

This paper presented several key findings for the semantic parsing research community. We first described an updated workflow for a state-of-the-art machine translation model to generate accurate and practical commands. We then introduced post-processing to replace placeholders in translated Bash Commands with the original parameters provided in the natural language. Our final contribution was a new dataset generated from scratch and an accompanying method for generating additional data. 

Our work provides essential foundations for building an automated system that translates natural language to Bash Commands. We were the first to (1) create an entirely valid Bash Command dataset from scratch and (2) provide a baseline accuracy of 31.6\% for translating natural language to Bash Commands on the new dataset. Our code is available in GitHub repository~\cite{BashGen2022}. The following is a summary of the lessons learned and gleaned from the research project:

\begin{itemize}
\item {\bf It is feasible to synthesize a large dataset of Bash Commands and corresponding English pair by adopting back-translation}. Generation from scratch is a major milestone and provides significant advantages over prior augmentation strategies. Our approach provided new opportunities for the generation of further natural language to computer code datasets and improved the effectiveness of Bash Command machine translation. 

\item {\bf To make translated commands practical they must be executable, therefore validity testing is important}. The conversion from a Bash Command template to an executable (or nearly executable) command shows both progress and promise of the usability and practicality of our translation pipeline. A more complete and streamlined process of converting natural language to valid, executable commands will become a larger focus as model accuracy continues to improve. 

\item {\bf It is necessary to create a hold-out dataset\footnote{A hold-out dataset is sourced differently from the original data, so it is more challenging compared to the test-set, which is sourced the same way as the training data and likely has the same distribution.} that is sourced in a different way to test model generality}. Although our dataset was six times larger than the original dataset and more diverse, the model exhibiting good performance on the test set failed to generalize to the original dataset and vice versa. As a result, the current datasets of Natural language to Bash Commands are still relatively small and insufficiently diverse to make robust models that generalize well to a hold-out set. This result could yield further developments and testing of different models to maximize performance on both our dataset and the original dataset.

\item {\bf ChatGPT has emerged as a potential solution for the natural language to Bash command (NLC2CMD) problem.} However, further research and experimentation are needed to fully explore the capabilities of ChatGPT and its potential as a solution for the NLC2CMD problem.

\end{itemize}

\bibliography{./bibliography/citation} 

\nocite{*} 
\section*{Biography}

\fbox{\parbox[t]{3cm}{\includegraphics[width=1.0\linewidth]{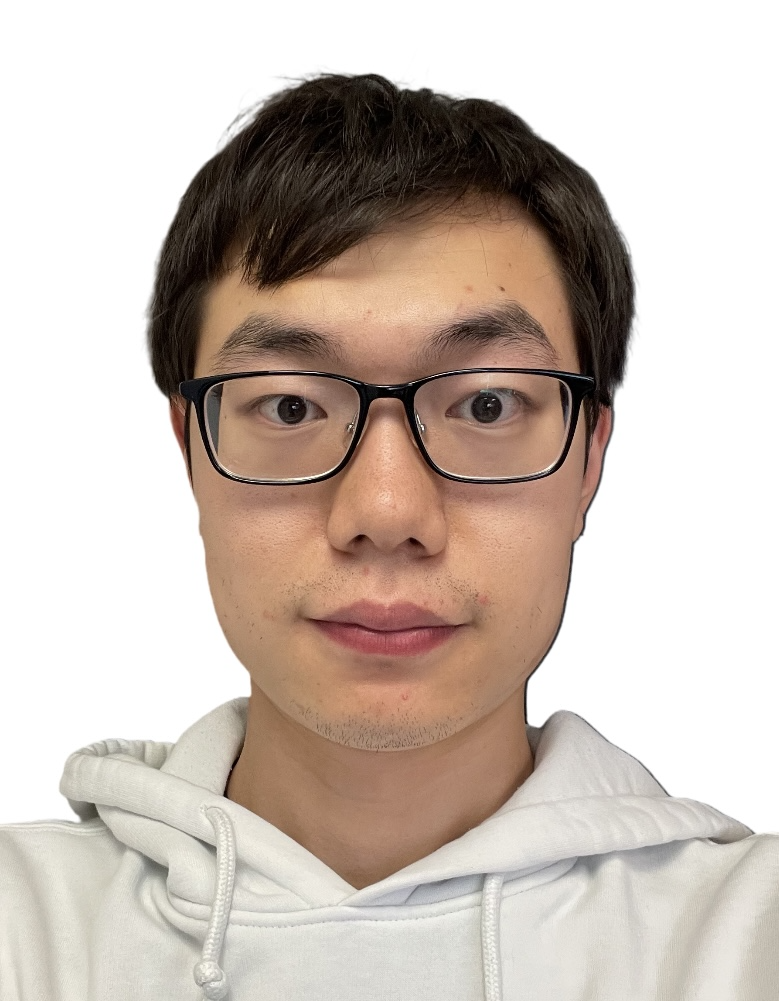}
}}

\medskip
\noindent
{\bf Quchen Fu}. Dr. Quchen Fu is an Applied Scientist in Microsoft's IC3-AI group. He holds a Ph.D. in Computer Science from Vanderbilt University in the Magnum research group under Dr. Jules White, specializing in NLP and Deep Learning. With a Master's degree from CMU, he served as a TA for courses like Cloud Computing and Cybersecurity. During his internships, he worked as a Backend Developer at Tencent. Notably, he developed a machine translation system that secured 1st place in the NLC2CMD competition at NeurIPS 2020. He also introduced FastAudio, a learnable audio frontend, which achieved a 27\% reduction in t-DCF for spoof speech detection. Additionally, he gained experience as a Deep Learning Software Engineer intern at Intel and as an Applied Scientist intern at Microsoft.

\fbox{\parbox[t]{3cm}{\includegraphics[width=1.0\linewidth]{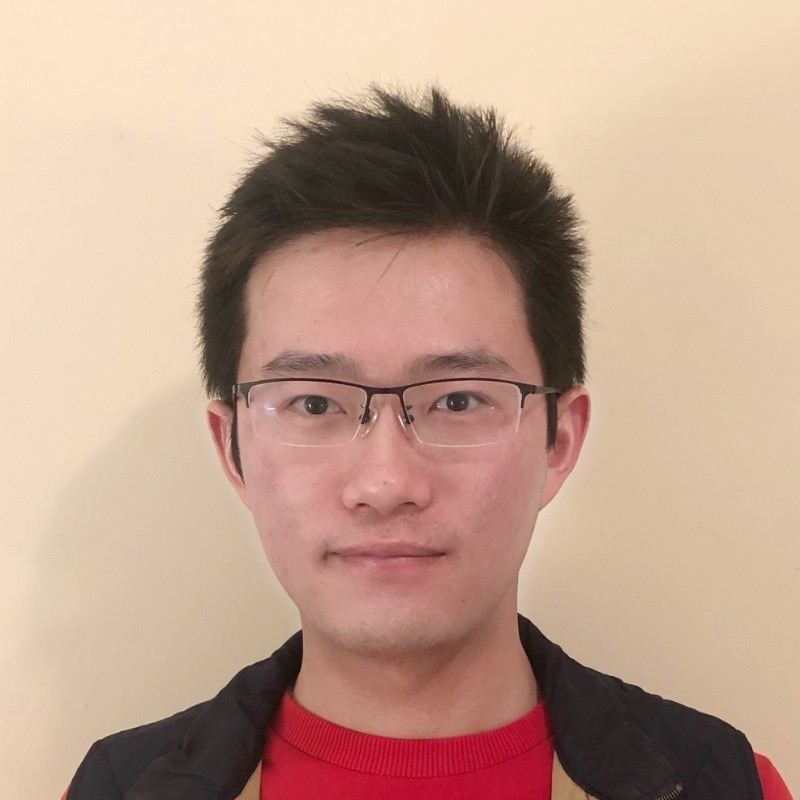}
}}

\medskip
\noindent
{\bf Zhongwei Teng}. Zhongwei Teng is pursuing a Ph.D. in Computer
Science in Vanderbilt University. His research interests
include speech verification, NLP and
machine learning.

\fbox{\parbox[t]{3cm}{\includegraphics[width=1.0\linewidth]{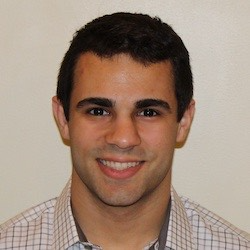}
}}

\medskip
\noindent
{\bf Marco Georgaklis}. Marco Georgaklis recently graduated from Vanderbilt University with a Bachelor’s in Computer Science. He will be starting as a Software Engineer at Google in the Fall of 2022.

\fbox{\parbox[t]{3cm}{\includegraphics[width=1.0\linewidth]{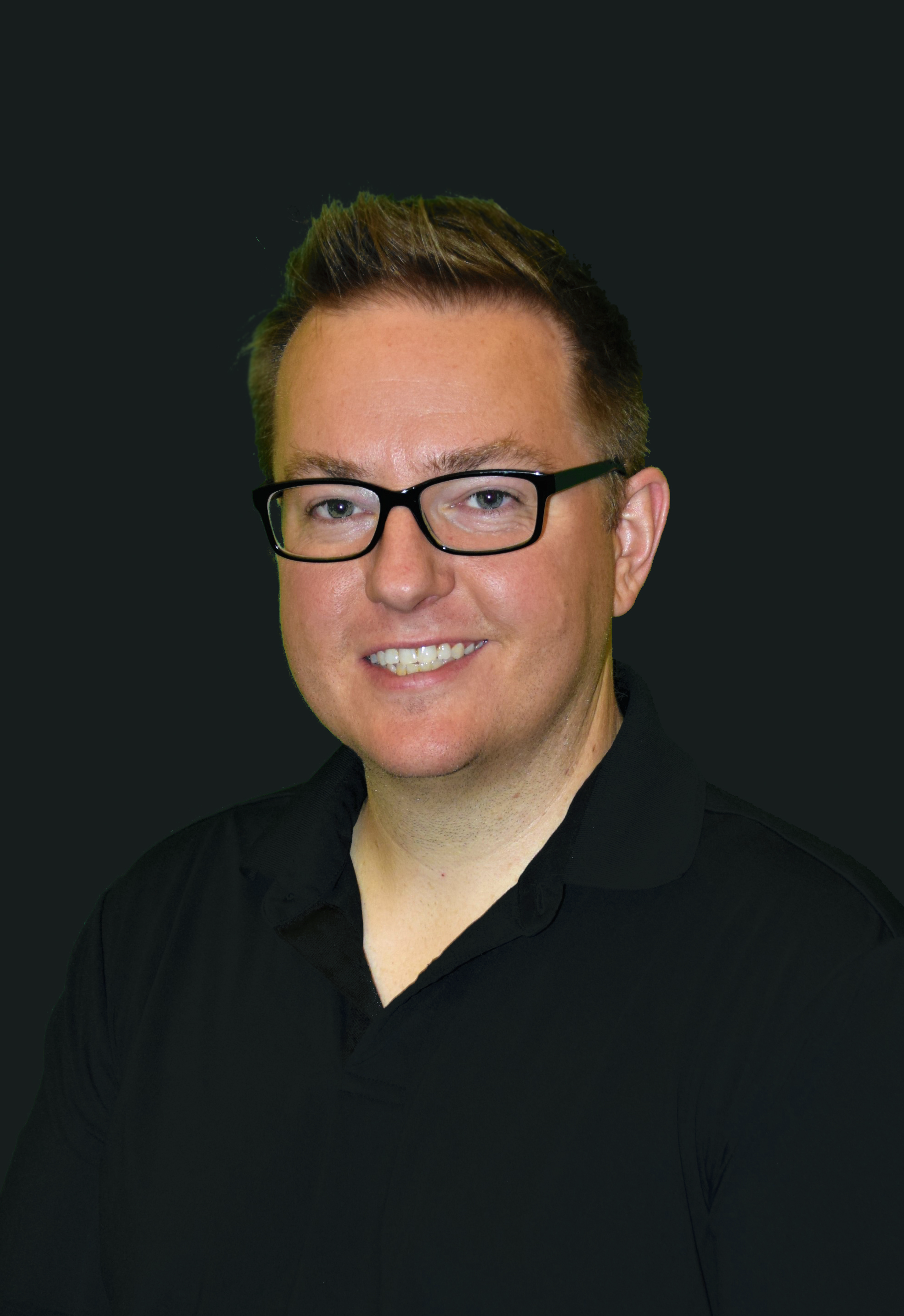}
}}

\medskip
\noindent
{\bf Jules White}. Dr. Jules White is Associate Dean of Strategic Learning Programs in the School of Engineering and Associate Professor of Computer Science in the Dept. of Computer Science at Vanderbilt University. He is a National Science Foundation CAREER Award recipient. His research has won multiple Best Paper Awards. He has also published over 150 papers. Dr. White’s research focuses on cyber-security and mobile/cloud computing in domains ranging from healthcare to manufacturing. His research has been licensed and transitioned to industry, where it won an Innovation Award at CES 2013, attended by over 150,000 people, was a finalist for the Technical Achievement at Award at SXSW Interactive, and was a top 3 for mobile in the Accelerator Awards at SXSW 2013. He has raised over \$12 million in venture backing for his startup companies. His research is conducted through the Mobile Application computinG, optimizatoN, and secUrity Methods (\href{http://www.magnum.io/people/jules.html}{MAGNUM}) Group at Vanderbilt University, which he directs.

\fbox{\parbox[t]{3cm}{\includegraphics[width=1.0\linewidth]{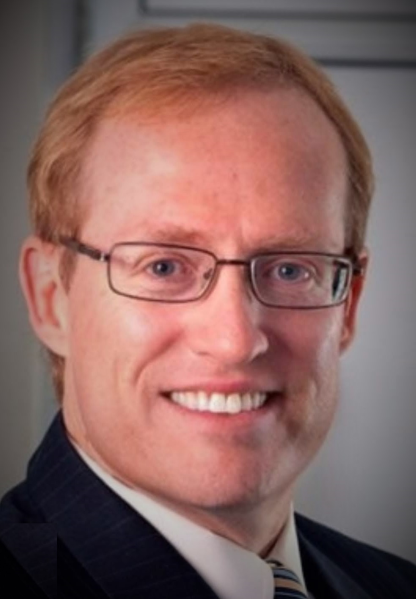}
}}

\medskip
\noindent
{\bf Douglas C Schmidt}. Dr. Douglas C. Schmidt is the Cornelius Vanderbilt Professor of Computer Science, Associate Chair of Computer Science, Co-Director at the Data Science Institute, and a Senior Researcher at the Institute for Software Integrated Systems, all at
Vanderbilt University. His research covers a range of software-related topics, including patterns, optimization techniques, and empirical analyses of frameworks and model-driven engineering tools that facilitate the development of mission-critical middleware for distributed real-time embedded (DRE) systems and intelligent mobile cloud computing applications. Dr. Schmidt received B.A. and M.A. degrees in Sociology from the College of William and Mary in Williamsburg, Virginia, and an M.S. and a Ph.D. in Computer Science from the University of California, Irvine (UCI) in 1984, 1986, 1990, and 1994, respectively.
\end{document}